\pdfoutput=1

\documentclass[11pt]{article}

\usepackage[final]{acl}

\usepackage{times}
\usepackage{latexsym}
\usepackage{hyperref}
\usepackage{placeins}
\usepackage{xcolor}
\definecolor{SoftGreen}{RGB}{0, 120, 0}  
\definecolor{SoftRed}{RGB}{180, 0, 0}     
\definecolor{SoftBlue}{RGB}{0, 0, 120}     

\usepackage{arydshln}     
\usepackage{makecell}     
\usepackage{array}        

\usepackage[T1]{fontenc}

\usepackage[utf8]{inputenc}

\usepackage{microtype}

\usepackage{inconsolata}

\usepackage{graphicx}
\usepackage{booktabs} 
\usepackage{multirow} 
\usepackage{amsmath}  
\usepackage{amssymb} 

\usepackage{algorithm}      
\usepackage{algorithmicx}   
\usepackage{algpseudocode}  

\newcommand{\rd}[1]{\left\lfloor #1 \right\rceil}  
\newcommand{\best}[1]{\textbf{#1}}

\title{Position IDs Matter: An Enhanced Position Layout for Efficient Context Compression in Large Language Models}

\author{
  \textbf{Runsong Zhao\textsuperscript{1,*}},
  \textbf{Xin Liu\textsuperscript{1,*}},
  \textbf{Xinyu Liu\textsuperscript{1}},
  \textbf{Pengcheng Huang\textsuperscript{1}}, \\
  \textbf{Chunyang Xiao},
  \textbf{Tong Xiao\textsuperscript{1,2,$\dagger$}} and \textbf{Jingbo Zhu\textsuperscript{1,2}}
  \\
  \textsuperscript{1} \normalsize{NLP Lab, School of Computer Science and Engineering, Northeastern University, Shenyang, China} \\
  \textsuperscript{2} \normalsize{NiuTrans Research, Shenyang, China} \\
  \normalsize{\{fhe43694, lxy1051493182, chunyangx\}@gmail.com,} \normalsize{lx\_meteor\_code@163.com} \\ \normalsize{pengcheng.neu@outlook.com,} \normalsize{\{xiaotong, zhujingbo\}@mail.neu.edu.com}
}

\begin{document}
\maketitle

\renewcommand{\thefootnote}{\fnsymbol{footnote}}
\footnotetext[1]{These authors contributed equally to this work.}
\footnotetext[2]{Corresponding author.}
\renewcommand{\thefootnote}{\arabic{footnote}}

\begin{abstract}


Using special tokens (e.g., gist, memory, or compressed tokens) to compress context information is a common practice for large language models (LLMs). However, existing approaches often neglect that position encodings inherently induce local inductive biases in models, causing the compression process to ignore holistic contextual dependencies.
We propose \textbf{Enhanced Position Layout (EPL)}, a simple yet effective method that improves the context compression capability of LLMs by only adjusting position IDs, the numerical identifiers that specify token positions.
EPL minimizes the distance between context tokens and their corresponding special tokens and at the same time maintains the sequence order in position IDs between context tokens, special tokens, and the subsequent tokens.
Integrating EPL into our best performing context compression model results in a 1.9 ROUGE-1 F1 improvement on out-of-domain question answering datasets on average.
When extended to multimodal scenarios, EPL leads to an average accuracy gain of 2.6 points for vision compression LLMs.~\footnote{Our code is available at \url{https://github.com/1azybug/EPL}.}

\end{abstract}

\section{Introduction}
In Transformer~\cite{vaswani2023attentionneed} architectures, special tokens have been widely adopted as compression carriers of contextual information across natural language processing~\cite{devlin-etal-2019-bert, liu2019robertarobustlyoptimizedbert,bulatov2022recurrent,ge2024incontext,li2024500xcompressorgeneralizedpromptcompression} and computer vision~\cite{dosovitskiy2021an,ye2025vocollamavisioncompressionlarge}. For context compression, the so-called soft prompt methods~\cite{chang2024efficientpromptingmethodslarge, li-etal-2025-prompt} employ encoders to condense long contexts into few special tokens, enabling decoders to perform inference based on compressed representations rather than raw inputs, thereby significantly reducing memory consumption and inference latency in long-context scenarios~\cite{jiang-etal-2024-hierarchical, xu2024conciseprecisecontextcompression, luo-etal-2025-beyond}. We illustrate typical soft prompt architectures in Figure~\ref{fig:main} where special tokens are appended at the end of the context, in order to capture the context semantics via causal attention mechanism in LLMs.

The design ensures full context visibility for special tokens. However, in Transformer architectures position IDs do not need to coincide with physical token positions and the model's perceived positional information is primarily determined by position IDs rather than physical token positions~\cite{vaswani2023attentionneed}. From this viewpoint, the local inductive biases introduced by position encodings~\cite{devlin-etal-2019-bert, vaswani2023attentionneed, su2023roformerenhancedtransformerrotary,JMLR:v21:20-074,press2022trainshorttestlong} weaken the efficacy of context compression under the default position layout (DPL), due to the substantial distance between the special tokens and the context tokens, as shown in Figure~\ref{fig:main} (DPL). In this paper, we carefully examine position layout designs and propose \textbf{Enhanced Position Layout (EPL)} for soft prompt architectures, which comprises \textbf{Uniform Position Layout (UPL)} and \textbf{Consistent Position Layout (CPL)}.

UPL redistributes special tokens' position IDs to achieve uniform distribution in the context tokens' position ID space, as exemplified in Figure~\ref{fig:main}. By uniformly assigning position IDs among context token position IDs, \textit{a priori}, most context tokens would have corresponding special tokens close to them. 
We assume that such a prior helps the special tokens compress the context. We formalize such intuitions, demonstrating the optimality of the UPL in Section~\ref{sec:UPL}. During compression, 
because special tokens are inserted and text chunks are reorganized, 
the position IDs between context, special tokens and subsequent tokens (e.g. reconstructed tokens or subsequent tokens such as QA pairs) can become inconsistent compared to their original positions before compression. Our proposed CPL in Section~\ref{sec:CPA} guarantees to maintain the order of the position ID sequence for different tokens, between different text chunks in their natural causal order.

We empirically apply EPL to two dominant context compression frameworks: ICAE~\cite{ge2024incontext} and  500xCompressor~\cite{li2024500xcompressorgeneralizedpromptcompression}. 
For the best model, on the autoencoding (AE) task, EPL yields a 1.8 BLEU gain and converges 9.7 times faster than DPL; on out-of-domain question answering (QA) tasks, EPL gives an average 1.9 ROUGE-1 F1 gain. To demonstrate the universality of EPL, we further conduct experiments by incorporating EPL into a more recent fine-grained compression framework~\cite{zhang2025long} and test our EPL equipped compressor on PwC dataset~\cite{ge2024incontext}, an instruction-tuning dataset containing diverse context-related prompts. In both scenarios, incorporating EPL brings consistent improvement.
When extending our application to multimodality with VoCo-LLaMA~\cite{ye2025vocollamavisioncompressionlarge}, EPL yields an average 2.6 accuracy gain on multimodal benchmarks. The improvement of EPL is consistent across base models of different scales. Further analysis
shows that both UPL (which aims for better context compression) and CPL (which maintains causal sequence ordering) are essential for the final performance improvement across tasks. Finally, our UPL attention map visualization confirms the usefulness of our specified prior:
UPL special tokens indeed focus more on tokens close to its assigned position IDs.

\section{Background}
\subsection{Local Bias of Position Encodings}
\label{sec:local_bias}

Transformer architectures~\cite{vaswani2023attentionneed} compute contextual token embeddings through position-invariant self-attention. Since natural language semantics crucially depend on token order, various position encodings (PEs) have been proposed to inject positional awareness including \textit{Sinusoidal PE}~\cite{vaswani2023attentionneed}, \textit{RoPE}~\cite{su2023roformerenhancedtransformerrotary}, \textit{Learnable PE}~\cite{devlin-etal-2019-bert}, \textit{T5 Bias}~\cite{JMLR:v21:20-074} and \textit{ALiBi}~\cite{press2022trainshorttestlong}, etc. All approaches share the inductive bias that adjacent tokens should correlate more strongly. Taking the PEs with the trigonometric encoding (e.g. Sinusoidal/RoPE) design as examples, the position embedding at a certain position is mostly similar to its neighbors and the similarity decays as the distance increases, see Appendix~\ref{sec:sin_pe} for more details. ALiBi enforces this inductive bias by applying distance-sensitive penalties to attention scores. 
We further show in Section~\ref{sec:learnable_local_bias} that Learnable PE and T5 Bias learn similar local bias through pre-training on natural text.


\begin{figure}[t]
\centering
\includegraphics[width=0.48\textwidth]{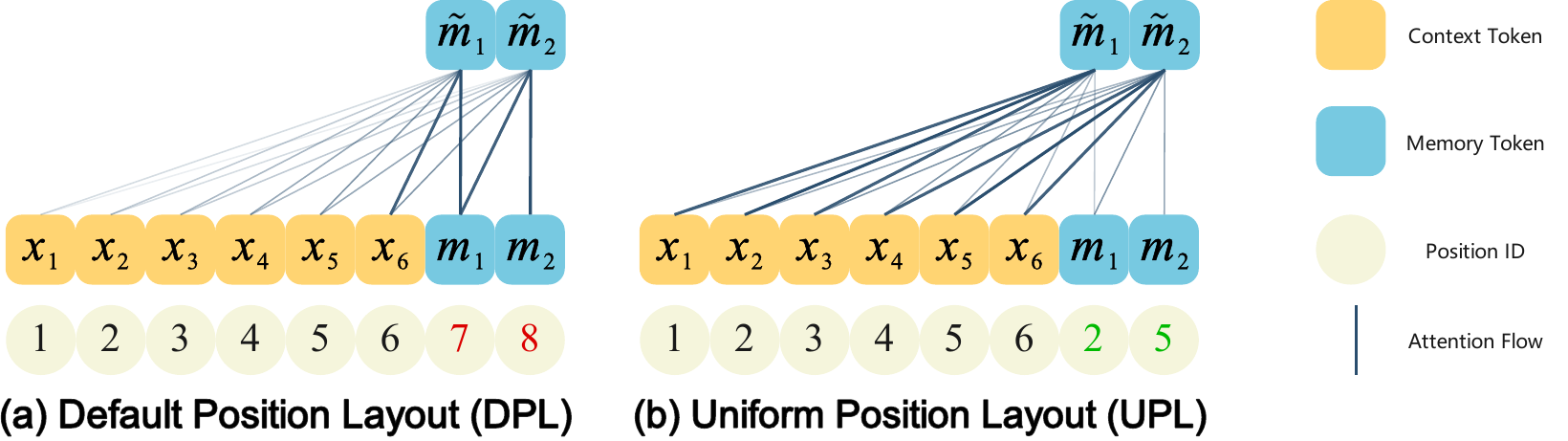}
\caption{
  Comparison of UPL and DPL.
  In prior work (DPL), memory tokens are assigned position IDs \textcolor{SoftRed}{7} and \textcolor{SoftRed}{8}. 
  Our method (UPL) allocates them to position IDs \textcolor{SoftGreen}{2} and \textcolor{SoftGreen}{5}.
Tokens with IDs in close proximity tend to exhibit higher attention scores.
}
\label{fig:main}
\end{figure}
\subsection{Position Layout}
While the local inductive bias for PEs is well known, less is known about the \textbf{position layout}. For any given token sequence and model, we refer to position layout as the actual position ID sequence that is assigned by the model to the token sequence. Notably, in this work, we remark that the local inductive bias applies to the position layout, not the physical token positions. 

This nuance is important but hardly noticeable because the \textbf{default position layout (DPL)} often coincides with the physical token positions as illustrated in DPL in Figure~\ref{fig:main} (see Table \ref{default-position-layout} for more details on DPL).
Such position layout is helpful for language modeling (LM) task since the task depends largely on its recent context~\cite{hu2024can, liu-etal-2024-forgetting}. 
However, as LLMs are becoming a ubiquitous tool, we hypothesize that careful position layouts for some tasks can inject helpful inductive bias. In the rest of the paper, we focus on LLM compression tasks as our testbed.

\section{Method}
\label{sec:method}

In Section~\ref{sec:soft_prompt}, we review existing LLM compression frameworks and their DPLs; in Section~\ref{sec:EPL} we describe our improved position layout.

\subsection{Soft Prompt Methods}
\label{sec:soft_prompt}

ICAE~\cite{ge2024incontext} is a widely used encoder-decoder soft prompt method. Its encoder compresses long context into a few memory tokens, after which the decoder performs inference conditioned only on the memory tokens to achieve faster inference speed. The left side of Figure~\ref{fig:icae} in the Appendix illustrates this process through an example. ICAE can be trained through two stages: continued pretraining and fine-tuning. Continued pretraining trains on a combination of AutoEncoding (AE) tasks and Language Modeling (LM) tasks, which trains the LLM encoder so that the encoded memory tokens enable a frozen LLM to losslessly reconstruct the original context and at the same time predict the subsequent tokens following the context to maintain ICAE's generation capability. The right side of Figure~\ref{fig:icae} illustrates the AE training process. Fine-tuning further trains the ICAE encoder to adapt to real-world applications such as question answering (QA) that we use in this work. We review the AE and LM pretraining as well as QA fine-tuning in the next subsections. To handle arbitrarily long contexts, we adopt the multi-chunk version of ICAE, which divides any long context into chunks for independent compression and then aggregates the resulting memory tokens to represent the complete long context. 500xCompressor~\cite{ge2024incontext} is similar to ICAE, with the difference of using the KV Cache of memory tokens as the compression carrier instead of the output of memory tokens.



\subsubsection{Pretraining}
\label{sec:pretraining_task}

During the pretraining stage, for a token sequence $X = \{x_0, x_1, \dots, x_{|X|-1}\}$, we take the first $p$ tokens as the context $X_{\text{context}} = \{x_0, x_1, \dots, x_{p-1}\}$ and the subsequent $|X|-p$ tokens as the completion $X_{\text{completion}} = \{x_{p}, x_{p+1}, \ldots, x_{|X|-1}\}$. The AE task only uses $X_{\text{context}}$, while the LM task leverages both $X_{\text{context}}$ and $X_{\text{completion}}$. 

\paragraph{Compress}
$X_{\text{context}}$ is partitioned into $k=\left\lceil p/L \right\rceil$ chunks (with chunk size $L$) 
where each chunk $S^{(i)} = \{x_{(i-1)L}, x_{(i-1)L+1}, \dots, x_{iL-1}\}$ is appended with a set of learnable memory tokens $M^{(i)} = \{m_0^{(i)}, m_1^{(i)}, \dots, m_{|M|-1}^{(i)}\}$. A LLM learns to encode each chunk into memory output tokens $\tilde{M}^{(i)}$ and key-value cache $KV^{(i)}$:
\begin{equation}
\tilde{M}^{(i)}, KV^{(i)} = \text{LLM}([S^{(i)}; M^{(i)}] \mid \theta_{\text{LoRA}})  
\label{eq:encoder}
\end{equation}
where $[;]$ denotes concatenation along the sequence dimension. All $M^{(i)}$ share the same learnable parameters $M$, and $\theta_{\text{LoRA}}$ denotes a set of low-rank adapter~\cite{hu2022lora} parameters for the LLM.
The final compressed representation is obtained by concatenating the results of each chunk: $ \tilde{M} = [\tilde{M}^{(1)}; \tilde{M}^{(2)}; \dots; \tilde{M}^{(k)}] $ or $ KV = [KV^{(1)}; KV^{(2)}; \dots; KV^{(k)}] $. 

\paragraph{Pretraining}
AE loss in ICAE is given by:
\begin{equation}
\mathcal{L}_\text{AE} = -\log P\big(X_{\text{context}} \mid [\tilde{M}; \texttt{[AE]}]\big)
\label{eq:AE}
\end{equation}
where $\texttt{[AE]}$ is a learnable token prompting the frozen decoder to generate $X_{\text{context}}$ as reconstruction.
Similar to the AE task, the loss for the LM task in ICAE is given by:
\begin{equation}
\mathcal{L}_\text{LM} = -\log P\big(X_{\text{completion}} \mid [\tilde{M}; \texttt{[LM]}]\big)
\label{eq:LM}
\end{equation}
where $\texttt{[LM]}$ is a learnable token that prompts the LLM to perform completion. We employ a weighted loss function for joint training\footnote{Unlike \citet{ge2024incontext}'s per-instance task allocation (AE with probability $\alpha$ /LM with $1-\alpha$), our joint training processes both tasks simultaneously through shared compressed representations, improving training efficiency by eliminating redundant context compression operations.} :
$$
\mathcal{L}_{\text{pretrain}} = \alpha \mathcal{L}_{\text{AE}} + (1-\alpha) \mathcal{L}_{\text{LM}}, \quad \alpha = 0.5
$$

\begin{table*}[h!]
  \centering
  \tiny
  \begin{tabular}{c|cccccccc|cccccccc}

    \toprule
    \multicolumn{17}{c}{$k=2, L=510, |M|=102, r=5, |X|=2040,p=1020$} \\    
    \midrule
    \noalign{\vspace{0.2em}}
    & \multicolumn{8}{c|}{$S^{(1)}$} & \multicolumn{8}{c}{$S^{(2)}$}\\
    \noalign{\vspace{0.2em}}
    \midrule
    \noalign{\vspace{0.2em}}
    & $x_{0}$ & $x_{1}$ & \dots & $x_{509}$ & $m_0^{(1)}$ & $m_1^{(1)}$ & \dots & $m_{101}^{(1)}$ &  $x_{510}$ & $x_{511}$ & \dots & $x_{1019}$ & $m_0^{(2)}$ & $m_1^{(2)}$ & \dots & $m_{101}^{(2)}$ \\
    \noalign{\vspace{0.2em}}
    \midrule
    \textbf{DPL(ICAE/500x)} & 0 & 1 & \dots & $509$ & \textcolor{SoftRed}{$510$} & \textcolor{SoftRed}{$511$} & \textcolor{SoftRed}{\dots} & \textcolor{SoftRed}{$611$} &  \textcolor{SoftRed}{0} & \textcolor{SoftRed}{1} & \textcolor{SoftRed}{\dots} & \textcolor{SoftRed}{$509$} & \textcolor{SoftRed}{$510$} & \textcolor{SoftRed}{$511$} & \textcolor{SoftRed}{\dots} & \textcolor{SoftRed}{$611$} \\
    \textbf{EPL} & \textcolor{SoftBlue}{$1$} & \textcolor{SoftBlue}{$2$} & \textcolor{SoftBlue}{\dots} & \textcolor{SoftBlue}{$510$} & \textcolor{SoftGreen}{$3$} & \textcolor{SoftGreen}{$8$} & \textcolor{SoftGreen}{\dots} & \textcolor{SoftGreen}{$508$} &  \textcolor{SoftBlue}{$511$} & \textcolor{SoftBlue}{$512$} & \textcolor{SoftBlue}{\dots} & \textcolor{SoftBlue}{$1020$} & \textcolor{SoftGreen}{$513$} & \textcolor{SoftGreen}{$518$} & \textcolor{SoftGreen}{\dots} & \textcolor{SoftGreen}{$1018$} \\
    \bottomrule

    
  \end{tabular}
  \caption{
    Position layout of the \textbf{encoder} for $S^{(1)}$ and $S^{(2)}$. 
    We show DPL's \textcolor{SoftRed}{problematic position IDs} in red, EPL's \textcolor{SoftGreen}{UPL-adjusted IDs} in green, and \textcolor{SoftBlue}{CPL-adjusted IDs} in blue.
    For more details, see Appendix~\ref{sec:position_layout_appendix}, 
    Table~\ref{tab:encoder_layout_chunk2_detail}.
}
  \label{tab:encoder_layout_chunk2}
\end{table*}

\begin{table*}[h!]
  \setlength{\tabcolsep}{7.5pt}
  \centering
  \tiny
  \begin{tabular}{c|ccccccccccc}

    \toprule

    \multicolumn{12}{c}{$k=2, L=510, |M|=102, r=5, |X|=2040,p=1020$} \\    
    
    \midrule
    \noalign{\vspace{0.2em}}
    & $\tilde{m}_{0}^{(1)}/KV_{0}^{(1)}$ & \dots & $\tilde{m}_{101}^{(1)}/KV_{101}^{(1)}$ & $\tilde{m}_{0}^{(2)}/KV_{0}^{(2)}$ & \dots  & $\tilde{m}_{101}^{(2)}/KV_{101}^{(2)}$ & $\texttt{[AE]}$ & $x_{0}$ & $x_{1}$ & \dots & $x_{1019}$ \\
    \noalign{\vspace{0.2em}}
    \midrule
    \textbf{DPL(ICAE)} & \textcolor{SoftRed}{$0$} & \textcolor{SoftRed}{\dots} & \textcolor{SoftRed}{$101$} & \textcolor{SoftRed}{$102$} & \textcolor{SoftRed}{\dots} & \textcolor{SoftRed}{$203$} & \textcolor{SoftRed}{$204$} & \textcolor{SoftRed}{$205$} & \textcolor{SoftRed}{$206$} & \textcolor{SoftRed}{\dots} & \textcolor{SoftRed}{$1224$} \\
    \textbf{DPL(500x)} & \textcolor{SoftRed}{$510$} & \textcolor{SoftRed}{\dots} & \textcolor{SoftRed}{$611$} & \textcolor{SoftRed}{$510$} & \textcolor{SoftRed}{\dots} & \textcolor{SoftRed}{$611$} & \textcolor{SoftRed}{$204$} & \textcolor{SoftRed}{$205$} & \textcolor{SoftRed}{$206$} & \textcolor{SoftRed}{\dots} & \textcolor{SoftRed}{$1224$} \\
    \textbf{EPL} & \textcolor{SoftGreen}{$3$} & \textcolor{SoftGreen}{\dots} & \textcolor{SoftGreen}{$508$} & \textcolor{SoftGreen}{$513$}  & \textcolor{SoftGreen}{\dots} & \textcolor{SoftGreen}{$1018$} & \textcolor{SoftBlue}{$0$} & \textcolor{SoftBlue}{$1$} & \textcolor{SoftBlue}{$2$} & \textcolor{SoftBlue}{\dots} & \textcolor{SoftBlue}{$1020$} \\
    \bottomrule

  \end{tabular}
  \caption{Position layout example of the \textbf{decoder} in AE Task.
    For more details, see Appendix~\ref{sec:position_layout_appendix}, 
    Table~\ref{tab:decoder_layout_ae_detail}.}
  \label{tab:decoder_layout_ae}
\end{table*}

\begin{table*}[h!]
\setlength{\tabcolsep}{6.8pt}
  \centering
  \tiny
  \begin{tabular}{c|ccccccccccc}
    \toprule

    \multicolumn{12}{c}{$k=2, L=510, |M|=102, r=5, |X|=2040,p=1020$} \\
    
    \midrule
    \noalign{\vspace{0.2em}}
     & $\tilde{m}_{0}^{(1)}/KV_{0}^{(1)}$ & \dots & $\tilde{m}_{101}^{(1)}/KV_{101}^{(1)}$ & $\tilde{m}_{0}^{(2)}/KV_{0}^{(2)}$ & \dots  & $\tilde{m}_{101}^{(2)}/KV_{101}^{(2)}$ & $\texttt{[LM]}$ & $x_{1020}$ & $x_{1021}$ & \dots & $x_{2039}$ \\
    \noalign{\vspace{0.2em}}
    \midrule
    \textbf{DPL(ICAE)} & \textcolor{SoftRed}{$0$} & \textcolor{SoftRed}{\dots} & \textcolor{SoftRed}{$101$} & \textcolor{SoftRed}{$102$} & \textcolor{SoftRed}{\dots} & \textcolor{SoftRed}{$203$} & \textcolor{SoftRed}{$204$} & \textcolor{SoftRed}{$205$} & \textcolor{SoftRed}{$206$} & \textcolor{SoftRed}{\dots} & \textcolor{SoftRed}{$1224$} \\
    \textbf{DPL(500x)} & \textcolor{SoftRed}{$510$} & \textcolor{SoftRed}{\dots} & \textcolor{SoftRed}{$611$} & \textcolor{SoftRed}{$510$} & \textcolor{SoftRed}{\dots} & \textcolor{SoftRed}{$611$} & \textcolor{SoftRed}{$204$} & \textcolor{SoftRed}{$205$} & \textcolor{SoftRed}{$206$} & \textcolor{SoftRed}{\dots} & \textcolor{SoftRed}{$1224$} \\    
    \textbf{EPL} & \textcolor{SoftGreen}{$3$} & \textcolor{SoftGreen}{\dots} & \textcolor{SoftGreen}{$508$} & \textcolor{SoftGreen}{$513$}  & \textcolor{SoftGreen}{\dots} & \textcolor{SoftGreen}{$1018$} & \textcolor{SoftBlue}{$1020$} & \textcolor{SoftBlue}{$1021$} & \textcolor{SoftBlue}{$1022$} & \textcolor{SoftBlue}{\dots} & \textcolor{SoftBlue}{$2040$} \\
    \bottomrule

  \end{tabular}
  \caption{Position layout example of the \textbf{decoder} in LM Task. 
    For more details, see Appendix~\ref{sec:position_layout_appendix}, 
    Table~\ref{tab:decoder_layout_lm_detail}.}
  \label{tab:decoder_layout_lm}
\end{table*}




\paragraph{Position Layout} Recall that for encoding, the token sequence starts with text chunk $S^{(i)}$ followed by memory tokens $M^{(i)}$. For decoding, the token sequence starts with memory token outputs $\tilde{M}$ for ICAE or $KV$ for 500xCompressor followed by $\texttt{[LM]}$ or $\texttt{[AE]}$ and then subsequent tokens (i.e. $X_{\text{context}}$ or $X_{\text{completion}}$).

The ICAE default position layout (DPL) always coincides with their physical token positions. For ICAE encoder for example, this means its DPL starts with position ID 0 and ranges till $|S^{(i)}| + |M^{(i)}| - 1$. For 500xCompressor, its encoder DPL also coincides with physical token positions. However, the position IDs of the decoder's $KV^{(i)}$ are the same as the position IDs of the encoder's $M^{(i)}$ (the memory tokens of the i-th chunk)\footnote{This is because the KV Cache has already cached the position information of the key\_state, see this \href{https://github.com/huggingface/transformers/blob/774dc274ac966f4bccbcd90d55bba23f6cca37ae/src/transformers/models/llama/modeling_llama.py\#L247-L252}{code snippet}.}, which implies that the decoder DPL for the $KV$\footnote{Recall that $KV$ concatenates all $KV^{(i)}$ with  $KV = [KV^{(1)}; KV^{(2)}; \dots; KV^{(k)}] $.} consists of $k$ repeated ranges from $|S^{(i)}|$ to $|S^{(i)}| + |M^{(i)}| - 1$. During decoding, DPL for the rest of the tokens (e.g. $\texttt{[AE]}, X_{\text{context}}$ in AE task) still coincides with their physical token positions. Table~\ref{tab:decoder_layout_ae} and Table~\ref{tab:decoder_layout_lm} show examples of ICAE and 500xCompressor DPL with $X_\text{context}$ having two chunks ($k$=2) under LM and AE task respectively. 

\subsubsection{Fine-tuning}
\label{sec:qa_task}
ICAE~\cite{ge2024incontext} allows further fine-tuning to enhance downstream task performance by enabling memory tokens to learn to focus on the tokens that are most related to the task; the training process is similar to LM pretraining.
Let each training instance consist of a triplet $(C, Q, A)$, where context $C$ is compressed into either $\tilde{M}$ (ICAE) or $KV$ (500xCompressor). The answer $A=\{a_0,a_1,...,a_{|A|-1}\}$ is generated conditioned on the compressed representation and the question $Q=\{q_0,q_1,...,q_{|Q|-1}\}$. The loss for QA task in ICAE is given by:
\begin{equation}
\label{eq:qa}
\mathcal{L}_\text{QA} = -\log P\big(A \mid [\tilde{M}; \texttt{[LM]}; Q]\big)
\end{equation}
The QA loss for 500xCompressor is similar to Eq.~\eqref{eq:qa}, with the difference of conditioning on $KV$ instead of $\tilde{M}$.

\paragraph{Position Layout}
The DPL for the QA task is similar to the DPL for the LM task and can be derived by replacing the $X_{\text{completion}}$ in the LM task with the concatenation $[Q;A]$. The resulting DPL for $[Q;A]$ coincides with their physical token positions. For the detailed DPL, see 
Table~\ref{tab:decoder_layout_qa} in the appendix.

\subsection{Enhanced Position Layout}
\label{sec:EPL}
For the DPL in soft prompt methods as described in Section~\ref{sec:soft_prompt}, we identify two limitations:
\paragraph{Distant Memory Tokens} Memory tokens in the soft prompt framework mainly aim to compress the context tokens so that the inference can be solely based on them to accelerate inference. However, the memory tokens in DPL consist of a continuous range (i.e. $510$-$611$ in Table~\ref{tab:encoder_layout_chunk2}) and are all very distant from the context token's range (i.e. $0$-$509$). Given the local inductive bias of PEs that we briefly review in Section~\ref{sec:local_bias},  it would be advantageous to have memory token position IDs to be both close to context token position IDs and covering the context token ID range. In Section~\ref{sec:UPL}, we propose Uniform Position Layout (UPL) that has memory token position layout covering the context token ID range while achieving minimum ID distances between memory tokens and context tokens.
\paragraph{Inconsistent Layout} Standard Transformer DPL coincides with physical token positions, which implies that tokens with larger position IDs follow the tokens with smaller position IDs, reflecting the causal relationship between the tokens through their assigned IDs.
However, we observe that some DPL does not comply with such properties. For example, 500xCompressor's decoder DPL as shown in Table~\ref{tab:decoder_layout_lm} starts with the $KV$ position IDs $\{510$, 511, \dots, $611\}$ but is followed by position ID sequence $\{204, 205, \dots, 1224\}$ representing $[\texttt{[LM]}; X_{\text{completion}}]$. In Section~\ref{sec:CPA}, we detail our Consistent Position Layout (CPL) design which guarantees the resulting position layout to maintain the causal structure amongst position IDs. We hypothesize that aligning the causal structures for position IDs
will benefit performance.

    

\subsubsection{Uniform Position Layout}
\label{sec:UPL}

\begin{algorithm}[t]
\caption{Generate Uniformly Distributed Compression Position IDs}
\label{alg:uniform_compression}
\begin{algorithmic}[1]
\Require Memory tokens count $|M|$, start/end IDs $v_1$, $v_L$
\Ensure Uniform positions $U^*$
\State $r \gets (v_L - v_1 + 1)/|M|$
\State $o \gets \frac{r-1}{2}$
\State $U^* \gets \text{torch.linspace}\bigl(v_1+o, 
           v_L-o,\;|M|\bigr)$
\State \Return $\text{torch.round}(U^*)$
\end{algorithmic}
\end{algorithm}


Recall that for memory token position layout, we aim to achieve two objectives: (1) the memory token position ID should be close to context tokens (2) for any context token, there is some memory token(s) whose IDs are close to it. Figure~\ref{fig:main} (b) \textbf{Uniform Position Layout (UPL)} illustrates such design, for any context token position ID, the nearest memory token position ID does not deviate more than 1, which sets contrast to DPL where the position ID of the first context token
$x_1$
is far from all memory token position IDs. In the following, we formalize the desiderata 
to derive analytically the optimal position layout UPL. 

Given a sequence of context token position IDs \( V = \{v_1, v_2, \dots, v_L\} \), \( v_i \in \mathbb{N} \) and  \( v_{i+1}-v_{i} = 1 \) $\forall i$, we aim to devise an algorithm to find position IDs for \( |M| \) memory tokens \( U = \{u_1, u_2, \dots, u_{|M|}\} \), \( u_j \in \mathbb{N} \)~\footnote{Although non-integer position IDs are valid in RoPE~\cite{su2023roformerenhancedtransformerrotary}, they have not been encountered during pre-training, making it difficult for the model to effectively utilize these non-integer position IDs.}, that minimize the following function:
\[
\max_{v_i \in V} \left( \min_{u_j \in U} |v_i - u_j| \right)
\]
where \( \min_{u_j \in U} |v_i - u_j| \) represents the distance from the \( i \)-th context token to its nearest memory token,
and \( \max_{v_i \in V }\) takes the maximum of all minimum distances across context tokens.

The optimal solution divides \(V\) evenly into \(|M|\) groups, with each group containing at most \(\left\lceil r \right\rceil\) tokens ($r=\frac{L}{|M|}$), and assigns each memory token the middle position of each group. In this case, the maximum distance from any context token to its nearest memory token is \(\left\lfloor \frac{\left\lceil r \right\rceil}{2} \right\rfloor\)\footnote{Note that any position layout will have its maximum distance \(\geq  \left\lfloor \frac{\left\lceil r \right\rceil}{2} \right\rfloor\), proving the optimality. }. Intuitively, the solution spreads memory token position IDs uniformly in the range of context token position IDs to ensure that no context token position ID is too far away. We detail the memory token position layout algorithm in Algorithm~\ref{alg:uniform_compression} that we apply for each chunk to be compressed. Table~\ref{tab:encoder_layout_chunk2} EPL row shows how the memory token position layout in UPL differs from  DPL.
\subsubsection{Consistent Position Layout}
\label{sec:CPA}
In this subsection, we propose consistent position layout (CPL) to ensure that the decoder position layout maintains the causal sequence order in position
IDs between context tokens, $\texttt{[LM]}$/$\texttt{[AE]}$, and
the subsequent tokens. As shown in Table ~\ref{tab:decoder_layout_ae} and \ref{tab:decoder_layout_lm} EPL rows, we keep memory token position layout unchanged compared to its encoding stage.

For tokens in $X_{\text{context}}$ and $X_{\text{completion}}$, we simply assign their original sequence positions as their position IDs.
For example, in the AE task, the token sequence $[$\texttt{[AE]}$; X_{\text{context}}]$ will be equipped with the position layout $\{0, 1, \dots, p\}$ where $p$ is the context length to reflect the tokens to be reconstructed from memory tokens.\footnote{Remark that the procedure is the inverse of memory token construction presented in Table~\ref{tab:encoder_layout_chunk2}. } For the LM task, the position layout is $\{p, p+1, \dots, |X|\}$ for $[$\texttt{[LM]}$; X_{\text{completion}}]$ as $X_{\text{completion}}$ logically follows $X_{\text{context}}$ in the physical token space. Table~\ref{tab:decoder_layout_ae} and Table~\ref{tab:decoder_layout_lm} EPL rows show concrete CPL during decoding through examples.

\section{Experimental Results}
\label{sec:experimental_results}

\subsection{Experimental Setup}
\label{subsec:experimental_setup}

\paragraph{Data}
For continued pretraining, we utilize the~\href{https://huggingface.co/datasets/DKYoon/SlimPajama-6B}{SlimPajama-6B}~\cite{cerebras2023slimpajama} corpus. To evaluate model fine-tuning performance, we use the \href{https://huggingface.co/datasets/mrqa-workshop/mrqa}{MRQA}~\cite{fisch2019mrqa} dataset as our testbed. The dataset contains evaluation on both in-domain scenarios where the validation dataset has its training counterpart used during training and out-of-domain scenarios. We report results from both settings but mainly discuss results for out-of-domain scenarios as it assesses more critically the soft prompt compression effectiveness.

\paragraph{Model Configuration}
We evaluate our method on \href{https://huggingface.co/meta-llama/Llama-3.2-1B}{Llama-3.2-1B}~\cite{grattafiori2024llama3herdmodels}. For efficient adaptation, we apply LoRA\cite{hu2022lora} to the query and value projection matrices within the multi-head attention layers of the encoder. The LoRA rank is set to 128, and the LoRA alpha is set to 256.  Following ICAE~\cite{ge2024incontext}, we do not train the decoder. In our default configuration, the number of memory tokens is $|M|=102$, the chunk size is $L=510$~\footnote{As context often exceeds the chunk size, we extensively evaluate multi-chunk settings, contrary to ICAE and 500xCompressor.}, implying a $r=5$ compression ratio. All models are further pretrained and fine-tuned for 20k steps with a batch size of 16. Further hyperparameter details can be found in Appendix~\ref{sec:hyperparamters}, Table~\ref{tab:hyperparamters}.



\subsection{Fine-tuning Results}
\label{sec:results}

Following ICAE and 500xCompressor, we pretrain then fine-tune Llama-3.2-1B; we integrate our EPL changes described in Section~\ref{sec:EPL} into different architectures respectively, then follow the same pretraining and fine-tuning steps. We evaluate the downstream performance using MRQA~\cite{fisch-etal-2019-mrqa} for different experiments. We assess the quality of the model's answers using ROUGE-1 F1~\cite{lin-2004-rouge} and Exact Match (EM) and report out-of-domain results in Table~\ref{tab:ood_results_main}.


\begin{table*}[h]
    \centering
    \tiny
    \begin{tabular}{l | c c |c c c c c c c c c c c c c c}
        \toprule
        & \textbf{LM} & \textbf{AE} &\multicolumn{2}{c}{\textbf{BioASQ}} & \multicolumn{2}{c}{\textbf{DROP}} & \multicolumn{2}{c}{\textbf{DuoRC}} & \multicolumn{2}{c}{\textbf{RACE}} & \multicolumn{2}{c}{\textbf{RE}} & \multicolumn{2}{c}{\textbf{TQA}} & \multicolumn{2}{c}{\textbf{Avg.}} \\
        \cmidrule(lr){2-2} \cmidrule(lr){3-3} \cmidrule(lr){4-5} \cmidrule(lr){6-7} \cmidrule(lr){8-9} \cmidrule(lr){10-11} \cmidrule(lr){12-13} \cmidrule(lr){14-15} \cmidrule(lr){16-17}
        & PPL & BLEU & F1 & EM & F1 & EM & F1 & EM & F1 & EM & F1 & EM & F1 & EM & F1 & EM \\
        \midrule
        ICAE(DPL) & 12.18 & 31.80 & 38.39 & 26.99 & \best{37.13} & \best{27.28} & 27.90 & 18.32 & 21.67 & 4.30 & \best{76.98} & \best{64.45} & 37.62 & 22.55 & 39.95 & 27.32 \\
        ICAE(EPL) & \best{11.42} & \best{95.98} & \best{42.66} & \best{29.19} & 37.03 & 26.95 & \best{32.50} & \best{21.39} & \best{26.87} & \best{5.34} & 76.90 & 64.31 & \best{47.25} & \best{29.81} & \best{43.87} & \best{29.50} \\
        500x(DPL) & 11.22 & 93.73 & 44.22 & 31.38 & 40.47 & 29.34 & 32.20 & 21.19 & 27.67 & 5.93 & \best{83.03} & \best{71.98} & 47.00 & 28.88 & 45.76 & 31.45 \\
        500x(EPL) & \best{10.80} & \best{98.50} & \best{46.11} & \best{31.91} & \best{40.94} & \best{29.41} & \best{39.07} & \best{26.32} & \best{31.89} & \best{7.27} & 80.39 & 67.71 & \best{49.80} & \best{31.14} & \best{48.03} & \best{32.29} \\
        \bottomrule
    \end{tabular}
    \caption{Pretraining and fine-tuning results. For the pretraining results, we report perplexity for completion 
    and BLEU-4~\cite{papineni-etal-2002-bleu} score for reconstruction 
    calculated on a held-out set of 1k examples. For the fine-tuning results, we report out-of-domain results, which include 6 datasets: BioASQ~\cite{Tsatsaronis2015}, DROP~\cite{dua-etal-2019-drop}, DuoRC~\cite{saha-etal-2018-duorc}, RACE~\cite{lai-etal-2017-race}, Relation Extraction (RE)~\cite{levy-etal-2017-zero}, and TextbookQA (TQA)~\cite{Kembhavi_2017_CVPR}. F1 in all the tables is ROUGE-1 F1~\cite{lin-2004-rouge}.
    }
    \label{tab:ood_results_main}
\end{table*}

For both ICAE and 500xCompressor, incorporating EPL significantly improves the performance. The average ROUGE-1 F1 improves from 39.95 to 43.87 for ICAE and improves from 45.76 to 48.03 for 500xCompressor. The improvement is observed for most domains, suggesting that the method is overall effective. We observe a similar improvement for in-domain settings (see Table~\ref{tab:id_results} in the appendix). 

To further validate the effectiveness of EPL for general context compression, we test on the PwC dataset~\cite{ge2023extensible} which contains diverse scenarios and question types. The results in Table~\ref{tab:pwc_results} show that EPL significantly improves downstream task performance compared to its counterpart, demonstrating its effectiveness.

\begin{table}[h]
\setlength{\tabcolsep}{12pt}
\centering
\small
\begin{tabular}{l c c}
\toprule
 & ROUGE-1 F1 & BLEU4 \\
\midrule
ICAE-1B (DPL) & 42.71 & 18.05 \\
ICAE-1B (EPL) & \best{45.59} & \best{21.04} \\
\bottomrule
\end{tabular}
\caption{Results on PwC~\cite{ge2024incontext}.}
\label{tab:pwc_results}
\end{table}

\subsection{Pretraining Results}
\label{sec:pretraining_results}
Through pretraining, LLMs have learned to compress context into memory tokens, allowing evaluation over memory tokens for their reconstruction and language modeling capability. The evaluation methodology is widely adopted for soft prompting~\cite{ge2024incontext, li2024llamavid} and we expect EPL to bring a similar improvement to the fine-tuning settings since EPL incorporates useful prior to reconstruction and language modeling through its position layouts. The reconstruction quality and language modeling capability are evaluated using BLEU-4~\cite{papineni-etal-2002-bleu} and perplexity (PPL), respectively\footnote{Reconstruction texts are generated via greedy search.}.

Table~\ref{tab:ood_results_main} confirms the EPL improvement. For both ICAE and 500xCompressor architectures, we observe better language modeling capability with lower perplexity as well as better reconstruction capability with higher BLEU. The improvement is more significant with the weaker ICAE model but significant for both architectures.



\subsection{Applications to Multimodal Models}
\label{sec:multimodal}
As EPL can be applied to all applications that compress context into special tokens, in this subsection, we showcase its application in multimodality. We follow VoCo-LLaMA~\cite{ye2025vocollamavisioncompressionlarge} for visual question-answering tasks. Given a triplet $(I, Q, A)$, the model encodes image $I$ into a sequence of 576 visual tokens $V^t = \{vt_0, vt_1, ..., vt_{575}\}$, and subsequently compresses $V^t$ into the KV values of \textbf{V}isi\textbf{o}n \textbf{Co}mpression (VoCo) tokens. VoCo-LLaMA adopts a single training stage akin to fine-tuning stage in 500xCompressor and employs a single-forward via an attention mask (see Figure~\ref{fig:voco_mask} for the attention mask detail) to prevent $Q$ and $A$ from directly accessing $V^t$. VoCo-LLaMA uses DPL and we follow its experimental setup to examine the effect of changing DPL to EPL. The position layout changes are illustrated on top of Figure~\ref{fig:voco_mask}. 

We evaluate VoCo-LLaMA with 128 VoCo tokens (i.e. 4.5x compression ratio) and report performance on multimodal benchmarks.
As shown in Table~\ref{tab:voco_result}, VoCo-LLaMA combined with EPL significantly outperforms both its DPL counterpart from our reproduction and the results reported by \citet{ye2025vocollamavisioncompressionlarge}. We observe improvement across all three tasks, validating the universality of EPL.






\begin{table}[h]
\setlength{\tabcolsep}{7pt}
\centering
\small
\begin{tabular}{ccccc}
\toprule
 & \multicolumn{1}{c}{SQA(OOD)} & \multicolumn{1}{c}{MMB(OOD)} & \multicolumn{1}{c}{VQAv2(ID)} \\
\midrule
DPL* & - & 61.0 & 76.9 \\
DPL & 64.9 & 59.9 & 76.9 \\
EPL & \textbf{66.5} & \textbf{64.6} & \textbf{78.4} \\
\bottomrule
\end{tabular}
\caption{Results of VoCo-LLaMA on SQA~\cite{lu2022learn}, MMB~\cite{MMBench}, and VQAv2~\cite{balanced_vqa_v2}. Results marked with * are from ~\cite{ye2025vocollamavisioncompressionlarge}.}
\label{tab:voco_result}
\end{table}

\subsection{Ablation Studies}
\label{sec:ablation_results}


\begin{table}[h]
\setlength{\tabcolsep}{3pt}
\centering
\small
\begin{tabular}{l c c c c}
\toprule
 & \multicolumn{1}{c}{LM} & \multicolumn{1}{c}{AE} & \multicolumn{2}{c}{Out-of-Domain} \\
\cmidrule(lr){2-2} \cmidrule(lr){3-3} \cmidrule(lr){4-5} 
 & PPL & BLEU & F1 & EM \\
\midrule
ICAE-1B & 12.18 & 31.80 & 39.95 & 27.31 \\
\textit{+UPL} & 11.50 & 72.35 & 43.67 & 29.11 \\
\textit{+UPL \& CPL (i.e. EPL)} & \best{11.42} & \best{95.98} & \best{43.87} & \best{29.50} \\
\specialrule{0em}{0.1pt}{1pt}
\cdashline{1-5}[2pt/2pt]
\specialrule{0em}{0.1pt}{2pt}
500x-1B & 11.22 & 93.73 & 45.76 & 31.45 \\
\textit{+UPL} & 10.94 & 95.86 & 46.83 & 31.64 \\
\textit{+UPL \& CPL (i.e. EPL)} & \best{10.80} & \best{98.50} & \best{48.03} & \best{32.29} \\
\bottomrule
\end{tabular}
\caption{Ablation study on EPL integration. \textit{+UPL} applies only UPL to encoder; \textit{UPL \& CPL (i.e. EPL)} applies both UPL and CPL. Reported values are averages. More ablation results are in Tables~\ref{tab:ablation_id_results} and Tables~\ref{tab:ablation_ood_results}.
}
\label{tab:ablation_results}
\end{table}



We conduct ablation studies to analyze the independent effects of UPL and CPL in EPL (see Section~\ref{sec:EPL} for the detailed description of UPL and CPL). LM perplexity and AE BLEU are measured under the same experimental settings as in Section~\ref{sec:pretraining_results} while out-of-domain performance is measured in the same way as in Section~\ref{sec:results}. We show the results in Table~\ref{tab:ablation_results} which confirm that:

\paragraph{UPL improves performance by injecting compression prior} For all testing cases (AE and LM during pretraining and MRQA out-of-domain performance for fine-tuning) over all architectures (ICAE and 500xCompressor), UPL improves performance compared to its counterparts. We note that the improvement is more significant for the weaker ICAE model. For example, the MRQA out-of-domain ROUGE-1 F1 improves by 3.72 points while the improvement is 1.07 points for 500xCompressor. The results confirm that by carefully designing the memory position layout for compression tasks, the memory tokens obtain useful compression inductive prior, which improves the compression efficiency (i.e. AE task performance) and consequently downstream task performance as shown by MRQA out-of-domain performance.


\paragraph{CPL improves performance by maintaining token sequential orders} Similarly, we observe consistent improvement across the board by incorporating CPL. We note that the performance improvement for CPL on top of UPL is more prominent for the stronger 500xCompressor model; MRQA out-of-domain ROUGE-1 F1 improves by 1.20 points while the improvement is only 0.20 for ICAE.  We think this is because ICAE DPL maintains the token sequential orders between $X_{\text{context}}$, $\texttt{[LM]}$, and $X_{\text{completion}}$ while 500xCompressor does not. As illustrated in Table~\ref{tab:decoder_layout_lm}, while the ICAE DPL coincides with physical token sequence positions, the 500xCompressor DPL exhibits $KV$ position IDs larger than the $\texttt{[LM]}$ token position ID (and some $X_{\text{completion}}$ position IDs), breaking the causal relationship between $\{X_{\text{context}}, \texttt{[LM]}, X_{\text{completion}}\}$ in the position layout.

\subsection{Scalability Results}
\label{sec:scale_results}

We conduct experiments at 3B and 8B scales from the same model family to verify method scalability. Table~\ref{tab:scaled_results} shows that EPL achieves performance improvements at all scales and the improvement does not attenuate with larger scale. For example, at 8B scale, the MRQA ROUGE-1 F1 increases by 14.03 points for ICAE and by 1.89 points for 500xCompressor, which is similar to the improvement observed in 1B case in Table~\ref{tab:ood_results_main}, holding promise for the method's effectiveness on large-scale language models. More detailed results on MRQA can be found in Table~\ref{tab:id_results} and~\ref{tab:ood_results}.


\begin{table}[h]
\setlength{\tabcolsep}{8pt}
\centering
\small
\begin{tabular}{l c c c c}
\toprule
 & \multicolumn{1}{c}{LM} & \multicolumn{1}{c}{AE} & \multicolumn{2}{c}{Out-of-Domain} \\
\cmidrule(lr){2-2} \cmidrule(lr){3-3} \cmidrule(lr){4-5}
 & PPL & BLEU & F1 & EM \\
 
\midrule
\multicolumn{5}{l}{\textit{Llama-3.2-3B}} \\

ICAE(DPL)  & 10.33 & 48.12 & 42.49 & 29.87 \\
ICAE(EPL) & \best{9.07} & \best{97.05} &  \best{55.03} & \best{38.54} \\
500x(DPL) & 9.44 & 96.86 & 51.37 & 36.58 \\
500x(EPL) & \best{8.68} & \best{99.34} & \best{57.71} & \best{40.49} \\

\specialrule{0em}{0.1pt}{1pt}
\cdashline{1-5}[2pt/2pt]
\specialrule{0em}{0.1pt}{2pt}
\multicolumn{5}{l}{\textit{Llama-3.1-8B}} \\

ICAE(DPL)  & 9.22 & 49.08 & 43.09 & 30.01 \\
ICAE(EPL) & \best{7.61} & \best{98.82} & \best{57.12} & \best{39.76} \\
500x(DPL) & 7.61 & 97.68 & 57.42 & 40.32 \\
500x(EPL) & \best{7.40} & \best{99.49} & \best{59.31} & \best{41.45} \\

\bottomrule
\end{tabular}
\caption{Results at 
\href{https://huggingface.co/meta-llama/Llama-3.2-3B}{Llama-3.2-3B}~\cite{grattafiori2024llama3herdmodels}, and \href{https://huggingface.co/meta-llama/Llama-3.1-8B}{Llama-3.1-8B}.
Reported values are averages.
}
\label{tab:scaled_results}
\end{table}

\subsection{Result of Fine-Grained Compression}
\label{sec:FGC}

To further validate the effectiveness of EPL, we conduct additional experiments on Fine-Grained Compression (FGC)~\cite{zhang2025long}. FGC shares a similar motivation with our UPL, as it divides the context into finer-grained units and interleaves memory tokens among them. Since FGC places memory tokens after these fine-grained units, EPL can be naturally applied to enhance its performance by setting each memory token's position ID to the middle position of the corresponding fine-grained unit. 
Given that FGC utilizes the KV cache as its compression carrier like the 500xCompressor, we incorporate FGC into the 500xCompressor architecture maintaining the same experimental setup as in Section~\ref{sec:results}.


\begin{table}[h]
\setlength{\tabcolsep}{7pt}
\centering
\small
\begin{tabular}{l c c c c}
\toprule
 & \multicolumn{1}{c}{LM} & \multicolumn{1}{c}{AE} & \multicolumn{2}{c}{Out-of-Domain} \\
\cmidrule(lr){2-2} \cmidrule(lr){3-3} \cmidrule(lr){4-5} 
 & PPL & BLEU & F1 & EM \\
\midrule
500x-1B & 11.22 & 93.73 & 45.76 & 31.45 \\
\textit{+EPL} & \best{10.80} & 98.50 & \best{48.03} & 32.29 \\
\textit{+FGC} & 11.03 & 94.49 & 45.52 & 30.40 \\
\textit{+FGC \& EPL} & \best{10.80} & \best{98.58} & 47.54 & \best{32.45} \\
\bottomrule
\end{tabular}
\caption{Results of fine-grained compression~\cite{zhang2025long}.
}
\label{tab:FGC_results}
\end{table}

\paragraph{EPL consistently enhances FGC performance} As shown in Table~\ref{tab:FGC_results}, applying EPL to the FGC yields consistent performance improvements across all evaluation metrics. These results demonstrate that EPL's position layout adjustments provide a generalizable enhancement to existing compression methods.





\subsection{Sensitivity Analysis for Compression Ratios}
\label{sec:sen_analy}

We conduct a sensitivity analysis of compression ratios to evaluate EPL's robustness with experimental settings aligned with Table~\ref{tab:ood_results_main}.

\begin{table}[h]
\setlength{\tabcolsep}{4pt}
\centering
\small
\begin{tabular}{l c c c c}
\toprule
 & \multicolumn{1}{c}{LM} & \multicolumn{1}{c}{AE} & \multicolumn{2}{c}{Out-of-Domain} \\
\cmidrule(lr){2-2} \cmidrule(lr){3-3} \cmidrule(lr){4-5} 
 & PPL & BLEU4 & F1 & EM \\
\midrule
ICAE-1B (DPL) \textit{5x} & 12.18 & 31.80 & 39.95 & 27.32 \\
ICAE-1B (EPL) \textit{5x} & \best{11.42} & \best{95.98} & \best{43.87} & \best{29.50} \\
\specialrule{0em}{0.1pt}{1pt}
\cdashline{1-5}[2pt/2pt]
\specialrule{0em}{0.1pt}{2pt}
ICAE-1B (DPL) \textit{15x} & \best{12.18} & 8.12 & 34.46 & 23.33 \\
ICAE-1B (EPL) \textit{15x} & 12.43 & \best{52.90} & \best{36.71} & \best{23.90} \\
\specialrule{0em}{0.1pt}{1pt}
\cdashline{1-5}[2pt/2pt]
\specialrule{0em}{0.1pt}{2pt}
ICAE-1B (DPL) \textit{31x} & \best{12.30} & 6.83 & \best{33.46} & \best{22.65} \\
ICAE-1B (EPL) \textit{31x} & 12.55 & \best{18.09} & 31.63 & 21.01 \\
\specialrule{0em}{0.1pt}{1pt}
\cdashline{1-5}[2pt/2pt]
\specialrule{0em}{0.1pt}{2pt}
ICAE-1B (DPL) \textit{51x} & \best{12.18} & 3.81 & 29.99 & 20.39 \\
ICAE-1B (EPL) \textit{51x} & 12.55 & \best{10.19} & \best{31.46} & \best{21.01} \\
\bottomrule
\end{tabular}
\caption{Results of different compression ratios.}
\label{tab:compression_sensitivity}
\end{table}

As shown in Table~\ref{tab:compression_sensitivity}, the results demonstrate that EPL significantly outperforms DPL across compression ratios up to 15x, encompassing the majority of practical application scenarios, although improvement attenuates as the compression ratio goes up. This performance range is particularly relevant, as lossless encoders typically fail to achieve compression ratios exceeding 10x~\cite{kuratov-etal-2025-cramming}. EPL performance improvement becomes less consistent at higher compression ratios, which we discuss further in the Limitation Section.

\section{Analysis}
\label{sec:analysis}

\subsection{Training Curve}
\label{training_curve}
During pretraining, we observe that the adoption of EPL significantly accelerates the convergence speed of the AE loss. In the 8B-500xCompressor setting, for example, EPL reduces the training steps required to achieve an AE loss of 0.01 from 9.7k to 1.0k steps, while effectively mitigating AE loss fluctuations. This suggests that the prior information that EPL incorporates is well-suited for AE tasks. For more discussion, see Appendix~\ref{sec:training_loss}.

\subsection{Attention Visualization}
\label{attn_vision}
To verify whether the performance improvement is indeed due to more suitable attention patterns for compression tasks idealized in Figure~\ref{fig:main}, we visualize the summed attention matrices of all attention heads in the second and final layers of the ICAE-3B model (Figure \ref{fig:layer_2_attn} and Figure \ref{fig:layer_last_attn}) after the model has been fine-tuned on MRQA tasks.

For the second layer, we observe that the memory token under \textbf{UPL} attends to its surrounding\footnote{In terms of position IDs, not physical token positions.} context tokens, forming a slope (central top of Figure~\ref{fig:layer_2_attn}), in contrast to \textbf{DPL} that only exhibits self-attention among memory tokens, showing that the attention adheres to our specified prior through EPL even after learning\footnote{The angle of the slope is around $11^\circ$ ($\arctan(1/\textbf{5})=11.3^\circ$ at a compression rate of \textbf{5}), which is consistent with our UPL conception illustrated in Figure~\ref{fig:main}. }. In the final layer (Figure~\ref{fig:layer_last_attn}), \textbf{UPL} maintains the slope pattern, while \textbf{DPL} overcomes the position encoding resistance and attends to distant context tokens through learning without a clear pattern.


\subsection{Learnable PEs}
\label{sec:learnable_local_bias}
Our results rely on the local bias of PEs that are explicitly defined in ~\cite{vaswani2023attentionneed,su2023roformerenhancedtransformerrotary,press2022trainshorttestlong}. We examine in this section empirically the properties of learnable position encodings~\cite{devlin-etal-2019-bert, JMLR:v21:20-074}.

Figure~\ref{fig:bert_pos} shows the cosine similarity of BERT's~\cite{devlin-etal-2019-bert} position embeddings  where we observe significantly higher similarity between adjacent position embeddings compared to distant tokens. Similarly, the bias values of T5 Bias~\cite{JMLR:v21:20-074} decay with increasing relative distance as shown in Figure~\ref{fig:t5_bias}, indicating that attention is stronger among close tokens.

BERT's \texttt{[CLS]} token (position ID 0) shows a special pattern as shown in Figure~\ref{fig:bert_cls_pos}: in contrast to other tokens, its cosine similarities are not higher with its adjacent tokens. Given that the \texttt{[CLS]} token behaves like a compressed memory token trained using the next sentence prediction, the phenomenon motivates our current work, suggesting that different priors should be given to special tokens for best performance.

\section{Related Work}
\label{related_work}
\paragraph{Soft Prompt Methods}
GIST~\cite{Mu2023LearningTC} trains LLMs with modified attention mechanisms (similar to Figure~\ref{fig:voco_mask}) to compress prompt information into a few gist tokens.
AutoCompressor~\cite{Chevalier2023AdaptingLM} trains an LLM to recursively compress long prompts by combining compressed tokens with new sub-prompts in each iteration, ultimately collecting all compressed tokens to form a compact representation. 
ICAE~\cite{ge2024incontext} introduces an AE task enabling LLMs to pre-train compression capabilities on large-scale corpora, requiring only minimal parameter tuning for the encoder while freezing the decoder. 
500xCompressor~\cite{li2024500xcompressorgeneralizedpromptcompression} builds upon ICAE by changing the information carrier from memory token outputs to memory tokens' KV values. 
UniICL~\cite{gao2024unifyingdemonstrationselectioncompression} and SelfCP~\cite{Gao2024SelfCPCO} freeze both the encoder and decoder and train only a connector module to transform the encoder's output memory tokens into decoder inputs.
VoCo-LLaMA~\cite{ye2025vocollamavisioncompressionlarge} is the first to use LLMs for compressing visual tokens. Similar to GIST, it compresses visual token information into VoCo tokens through modified attention masks, outperforming methods like Q-Former~\cite{li2023blip2} and average pooling with linear projection~\cite{li2024llamavid}. None of these methods discuss the impact of position layout and our EPL can be applied to all these soft prompt methods.

\paragraph{Position Layout}
Although we are not aware of position layout work in the compression domain, we find related work in multimodality by considering memory tokens as another modality.
Due to images being two-dimensional and text being one-dimensional, handling mixed image-text positional layout remains an open question. Current mainstream approaches in multimodality flatten 2D images into 1D sequences~\cite{liu2023llava, liu2023improvedllava, fuyu-8b, lu2024deepseekvl, Sun_2024_CVPR}. Seeking a more elegant solution, \citet{kexuefm-10040} proposes RoPE-Tie by placing visual and text tokens along the diagonal $(y=x)$ in 2D space. Although \citet{kexuefm-10040} does not thoroughly validate its design,
as the approach maintains the sequence order between modalities and the internal locality of images and text respectively,
our empirical results suggest that the design can result in better performance. Qwen2-VL~\cite{Qwen2-VL} adopts similar designs in video domains.

\section{Conclusion}
\label{sec:conclusion}

We examine position layout, an understudied topic in context compression, and propose EPL for soft prompt methods. EPL improves over default position layout by bringing memory tokens close to its context and at the same time maintains the logical sequence token ordering among context, memory tokens and the subsequent tokens. Extensive experiments show EPL enhances compression efficiency and downstream performance across architectures and modalities.


Given LLM's widespread use, we believe examining position layout will benefit problems beyond context compression by adopting a prior tailored to tasks that can be expressed in the position layout. We hope EPL's success fosters research in this underexplored area.




\section*{Limitations}
\label{sec:limitation}

Throughout our experiments, we have tested and confirmed EPL's effectiveness under 5x, 15x, 51x compression ratios for text reconstruction and QA tasks and 4.5x compression ratio for visual QA. There remains a question as to whether our method is still effective at high compression ratio. We don't have a positive outlook on this question. 

First, EPL, requiring memory tokens to achieve uniform coverage across the context, intrinsically aligns with lossless compression scenarios (i.e., autoencoding tasks) which assume that ``all tokens in the context are equally important''; however, under high compression ratio (lossy compression), with information carrier capacity being limited, focus should be placed on important tokens, which violates our ``equally important'' assumption. Empirically, Figure~\ref{fig:bert_cls_pos} shows that the BERT's \texttt{[CLS]} would have higher attention to some tokens without a clear pattern. Our compression ratio analysis in Table~\ref{tab:compression_sensitivity} and results in Table~\ref{tab:voco_whole_result} on VoCo-LLaMA also show that EPL does not demonstrate any significant gains when used under high compression ratios. 

The analysis suggests that high-ratio lossy compression may require a compression mechanism that goes beyond our current work. At a higher level, this suggests that when one adapts position layout methods to different application scenarios, the success can highly depend on whether the conceived layout captures the underlying prior characteristics of specific tasks.


\section*{Acknowledgments}
This work was supported in part by the National Science Foundation of China (Nos. 62276056 and U24A20334), the Yunnan Fundamental Research Projects (No.202401BC070021), the Yunnan Science and Technology Major Project (No. 202502AD080014), and the Program of Introducing Talents of Discipline to Universities, Plan 111 (No.B16009).

\bibliography{main}

\clearpage

\appendix

\section{Detailed Position Layout}
\label{sec:position_layout_appendix}

Detailed position layouts are provided in this section. Table~\ref{default-position-layout} illustrates the position layout for LLMs without compression under standard conditions, reflecting natural language priors. Table~\ref{tab:encoder_layout_chunk2_detail} details the encoder's position layout. For the decoder, position layouts are described in Tables~\ref{tab:decoder_layout_ae_detail}, \ref{tab:decoder_layout_lm_detail}, and \ref{tab:decoder_layout_qa}, with slight variations depending on the specific task (AE, LM, or QA). Each of these tables (Tables~\ref{tab:encoder_layout_chunk2_detail}, \ref{tab:decoder_layout_ae_detail}, \ref{tab:decoder_layout_lm_detail}, and \ref{tab:decoder_layout_qa}) includes a formula at the top representing the generalized layout. Below the formula, an illustrative example is given using a context length $|C|=p=1020$, chunk size $L=510$, compression ratio $r=5$, total sequence length $|X|=2040$, question length $|Q|=50$, and answer length $|A|=5$.



\section{Overview of ICAE}
\label{sec:soft_prompt_methods}

\begin{figure*}[htbp!]
\centering
\includegraphics[width=0.98\textwidth]{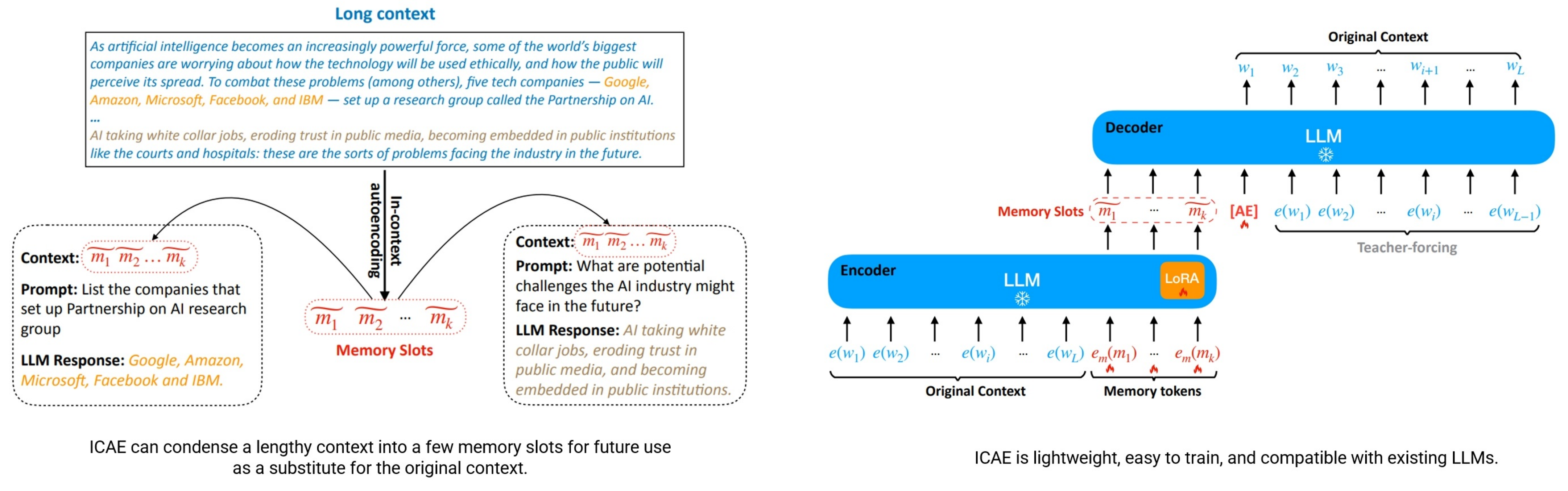}
\caption{Overview of the ICAE framework proposed by \cite{ge2024incontext}. Source: ICAE official \href{https://github.com/getao/icae}{repository} (CC0-1.0).}
\label{fig:icae}
\end{figure*}

ICAE~\cite{ge2024incontext} is an autoencoder framework to compress long contexts into short compact memory slots. 
The method operates by concatenating designated memory tokens to the end of the input sequence before an encoder processes the entire combined sequence. Subsequently, a decoder reconstructs the original sequence using only the information contained within the memory tokens. ICAE is trained in two main phases. It is first pretrained on massive text data using a combination of autoencoding and language modeling objectives, enabling it to generate memory slots that represent the original context. Following pretraining, the model is fine-tuned on instruction data for the purpose of producing desirable responses to various prompts. 
An overview of the ICAE framework is shown in Figure~\ref{fig:icae}.


\section{Hyperparameters}
\label{sec:hyperparamters}
For the 1B and 3B models, we perform continued pre-training on sequences with lengths $|X|$ ranging from 510 to 2040. For the 8B model, the input sequence length $|X|$ ranges from 510 to 4080 during continued pre-training. We take the first $p=\left\lfloor |X|/2 \right\rfloor$ tokens as the context.
Additional hyperparameters are listed in Table~\ref{tab:hyperparamters}.

\begin{table}[htbp!]
\centering
\small
\begin{tabular}{ll}
\toprule
\textbf{Hyperparameter} & \textbf{Value} \\
\midrule
Optimizer              & AdamW \\
Betas & (0.9, 0.95) \\
Weight decay & 0.1 \\
Learning rate          & 1e-4 (pretrain) \\
& 5e-5 (fine-tuning) \\
Scheduler & Constant \\
Batch size             & 16 \\
Warmup                 & 300 \\
Training steps              & 20k (pretrain) \\
& 20k (fine-tuning) \\
Clip norm              & 2.0 \\
\bottomrule
\end{tabular}
\caption{Hyperparameters for training}
\label{tab:hyperparamters}
\end{table}


\section{Detailed Results}
\label{sec:Addition_results}

In this section, we provide detailed results. 
Considering the potential risk of data leakage where LLMs may have encountered context information from evaluation datasets during the pretraining phase\cite{li2024500xcompressorgeneralizedpromptcompression}, we also report on \textbf{NoContext} and \textbf{FullContext} settings where \textbf{NoContext} performs inference solely based on the question and \textbf{FullContext} utilize the complete context and the question for inference. In both cases, we only train \texttt{[AE]} and \texttt{[LM]} tokens to guide the model executing corresponding tasks.

Table~\ref{tab:pretraining_results} fully presents the performance of ICAE and 500xCompressor of different scales on LM tasks and AE tasks. Table~\ref{tab:id_results} and Table~\ref{tab:ood_results} respectively show all results of these models in-domain and out-of-domain in MRQA. 
Ablation results in-domain and out-of-domain in MRQA are presented in Table~\ref{tab:ablation_id_results} and Table~\ref{tab:ablation_ood_results}, respectively.
Table~\ref{tab:voco_whole_result} shows all results of VoCo-LLaMA on multimodal benchmarks.

\begin{table}[htbp!]
    \centering
    \tiny
    \begin{tabular}{l c c c}
    \toprule
    & \textbf{PPL(AE)} & \textbf{PPL(LM)} & \textbf{BLEU(AE)} \\
    \midrule
    \multicolumn{4}{l}{Llama-3.2-1B} \\
    \midrule
    \textbf{NoContext} & \textcolor{SoftRed}{11.56} & \textcolor{SoftRed}{13.25} & \textcolor{SoftRed}{0.00} \\
    ICAE(DPL) & 1.40 & 12.18 & 31.80 \\
    ICAE(EPL) & \best{1.04} & \best{11.42} & \best{95.98} \\
    500x(DPL) & 1.04 & 11.22 & 93.73 \\
    500x(EPL) & \best{1.01} & \best{10.80} & \best{98.50} \\
    \textbf{FullContext} & \textcolor{SoftGreen}{1.02} & \textcolor{SoftGreen}{9.90} & \textcolor{SoftGreen}{33.49} \\
    \midrule
    \multicolumn{4}{l}{Llama-3.2-3B} \\
    \midrule
    \textbf{NoContext} & \textcolor{SoftRed}{9.58} & \textcolor{SoftRed}{11.02} & \textcolor{SoftRed}{0.00} \\
    ICAE(DPL) & 1.49 & 10.33 & 48.12 \\
    ICAE(EPL) & \best{1.02} & \best{9.07} & \best{97.05} \\
    500x(DPL) & 1.02 & 9.44 & 96.86 \\
    500x(EPL) & \best{1.00} & \best{8.68} & \best{99.34} \\
    \textbf{FullContext} & \textcolor{SoftGreen}{1.06} & \textcolor{SoftGreen}{8.25} & \textcolor{SoftGreen}{60.16} \\
    \midrule
    \multicolumn{4}{l}{Llama-3.1-8B} \\
    \midrule
    \textbf{NoContext} & \textcolor{SoftRed}{7.79} & \textcolor{SoftRed}{9.02} & \textcolor{SoftRed}{0.00} \\
    ICAE(DPL) & 1.58 & 9.22 & 49.08 \\
    ICAE(EPL) & \best{1.00} & \best{7.61} & \best{98.82} \\
    500x(DPL) & 1.01 & 7.61 & 97.68 \\
    500x(EPL) & \best{1.00} & \best{7.40} & \best{99.49} \\
    \textbf{FullContext} & \textcolor{SoftGreen}{1.00} & \textcolor{SoftGreen}{7.30} & \textcolor{SoftGreen}{98.00} \\
    \bottomrule
    \end{tabular}
    \caption{Perplexity of $X_{\text{context}}$ and $X_{\text{completion}}$, and BLEU score for the reconstruction quality of $X_{\text{context}}$, calculated on a held-out set of 1k examples.}
 \label{tab:pretraining_results}
\end{table}


\begin{table}[htbp!]
\centering
\tiny
\begin{tabular}{ccccc}
\toprule
 & \multicolumn{1}{c}{voco\_num} & \multicolumn{1}{c}{SQA(OOD)} & \multicolumn{1}{c}{MMB(OOD)} & \multicolumn{1}{c}{VQAv2(ID)} \\
\midrule
\textcolor{SoftRed}{Lower Bound*} & \textcolor{SoftRed}{1} & \textcolor{SoftRed}{60.7} & \textcolor{SoftRed}{22.3} & \textcolor{SoftRed}{41.2} \\
\midrule
DPL* & 2 & - & 60.1 & 73.5 \\
DPL & 2 & 66.1 & \textbf{61.5} & \textbf{73.7} \\
EPL & 2 & \textbf{67.6} & 61.4 & 69.5 \\
\midrule
DPL* & 128 & - & 61.0 & 76.9 \\
DPL & 128 & 64.9 & 59.9 & 76.9 \\
EPL & 128 & \textbf{66.5} & \textbf{64.6} & \textbf{78.4} \\
\textcolor{SoftGreen}{Upper Bound*} & \textcolor{SoftGreen}{576} & \textcolor{SoftGreen}{66.5} & \textcolor{SoftGreen}{64.0} & \textcolor{SoftGreen}{77.7} \\
\bottomrule
\end{tabular}
\caption{Results of VoCo-LLaMA on SQA~\cite{lu2022learn}, MMB~\cite{MMBench}, and VQAv2~\cite{balanced_vqa_v2} from our experiments. Results marked with * are from ~\cite{ye2025vocollamavisioncompressionlarge}.}
\label{tab:voco_whole_result}
\end{table}


\begin{figure}[htbp!]
\centering
\includegraphics[width=0.42\textwidth]{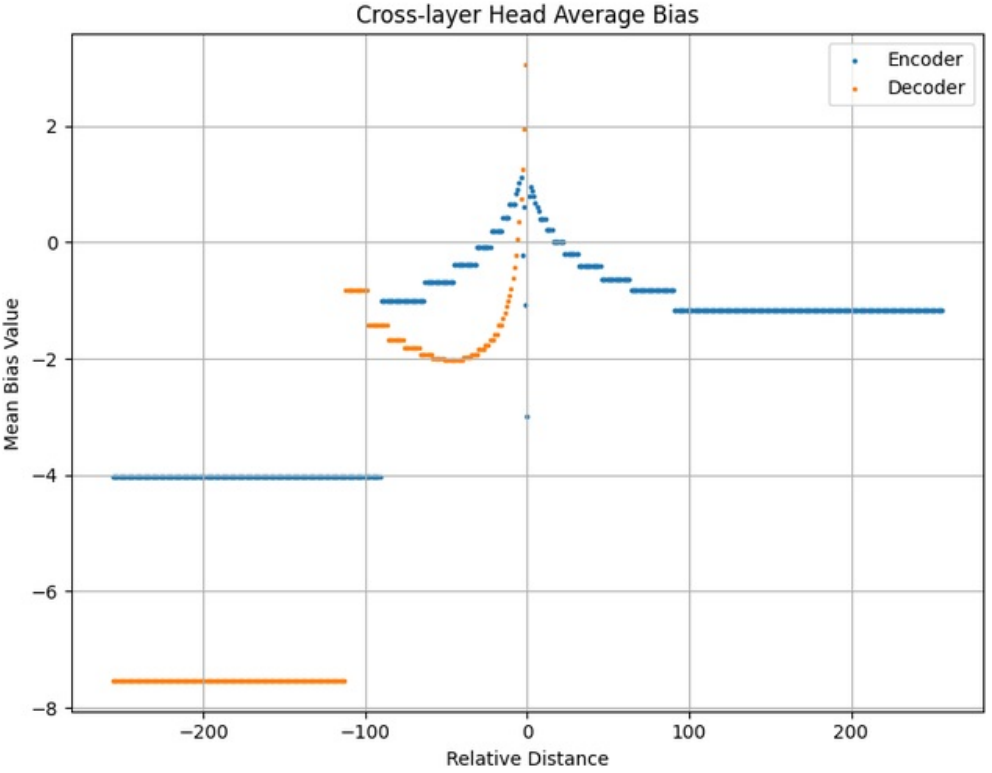}
\caption{T5 bias.
}
\label{fig:t5_bias}
\end{figure}

\section{Training Curves}
\label{sec:training_loss}
\begin{figure*}[htbp!]
\centering
\includegraphics[width=0.98\textwidth]{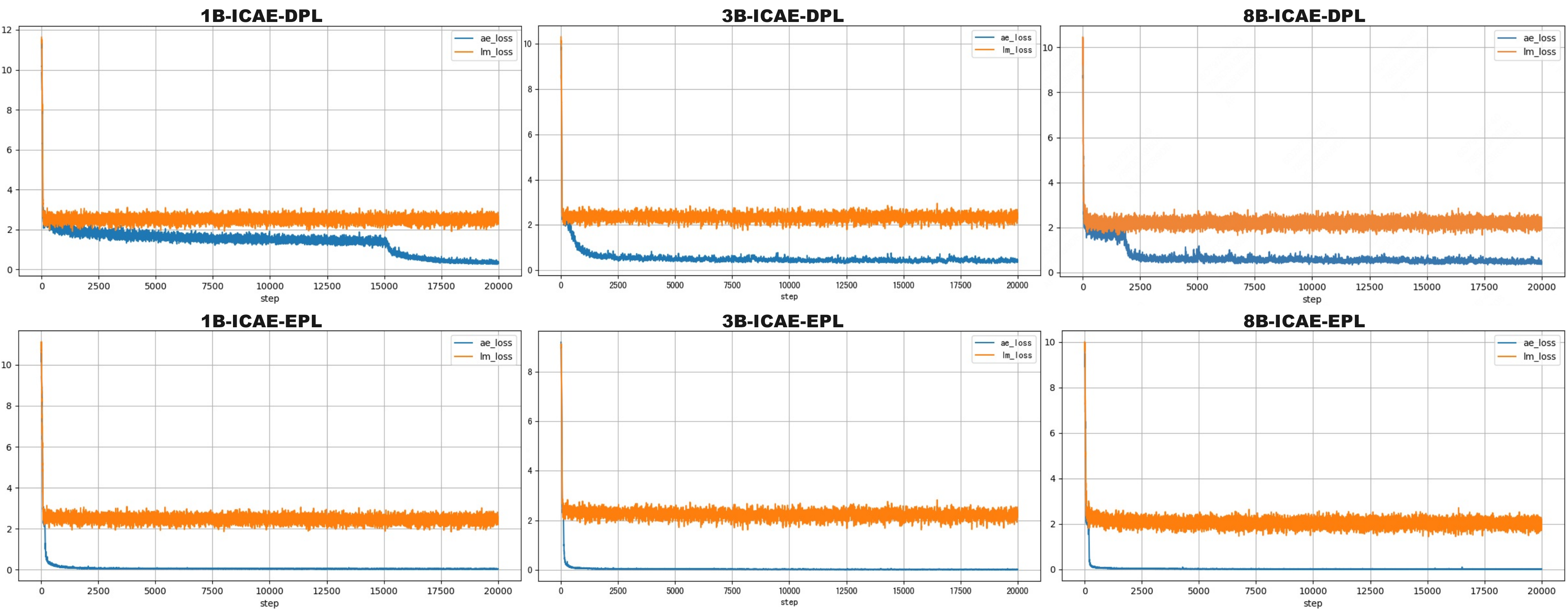}
\caption{Training Loss of ICAE.}
\label{fig:icae_loss}
\end{figure*}

\begin{figure*}[htbp!]
\centering
\includegraphics[width=0.98\textwidth]{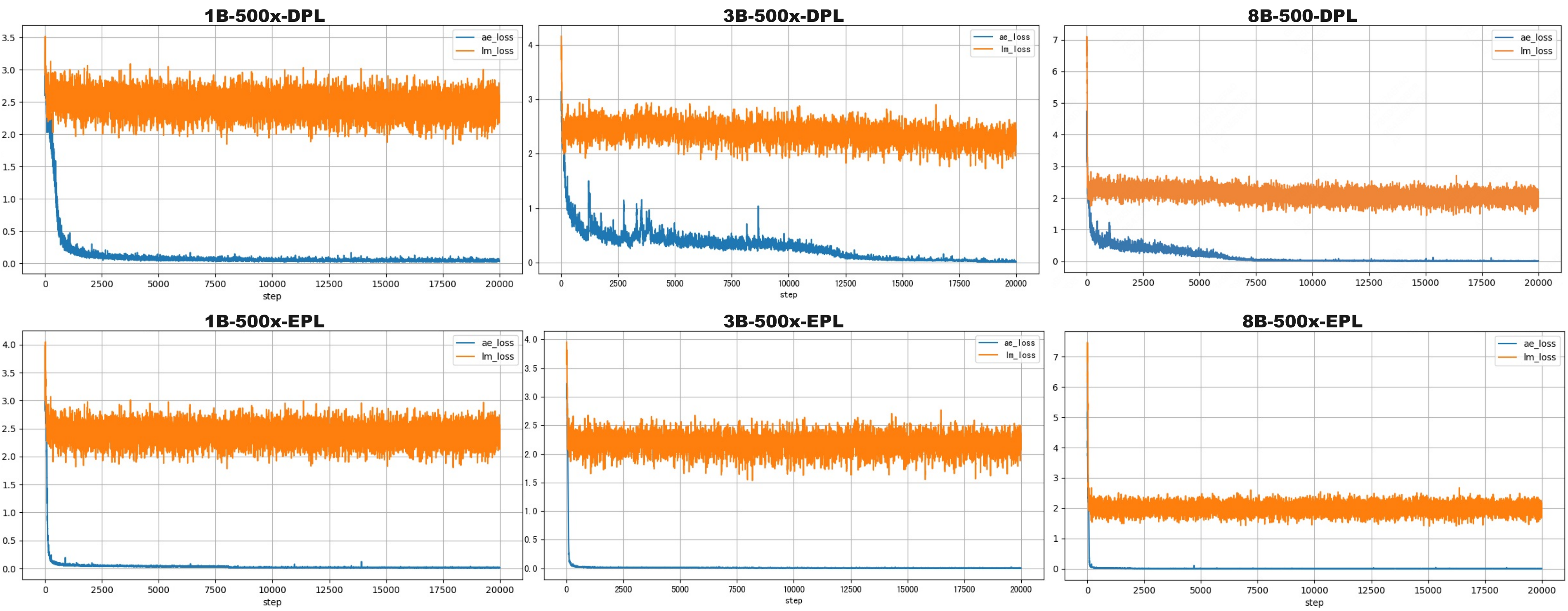}
\caption{Training Loss of 500xCompressor.}
\label{fig:500x_loss}
\end{figure*}

This section provides training curves. Figure~\ref{fig:icae_loss} and Figure~\ref{fig:500x_loss} respectively show the training curves of ICAE and 500xCompressor of different scales during pretraining. For AE loss, it can be observed that ICAE(DPL) struggles to decrease to 0, while 500xCompressor(DPL) requires a period of oscillation before converging near 0. When UPL is applied, their AE loss rapidly converges to around 0.

\section{VoCo-LLaMA}
\label{sec:voco_llama}
\begin{figure*}[htbp!]
\centering
\includegraphics[width=0.78\textwidth]{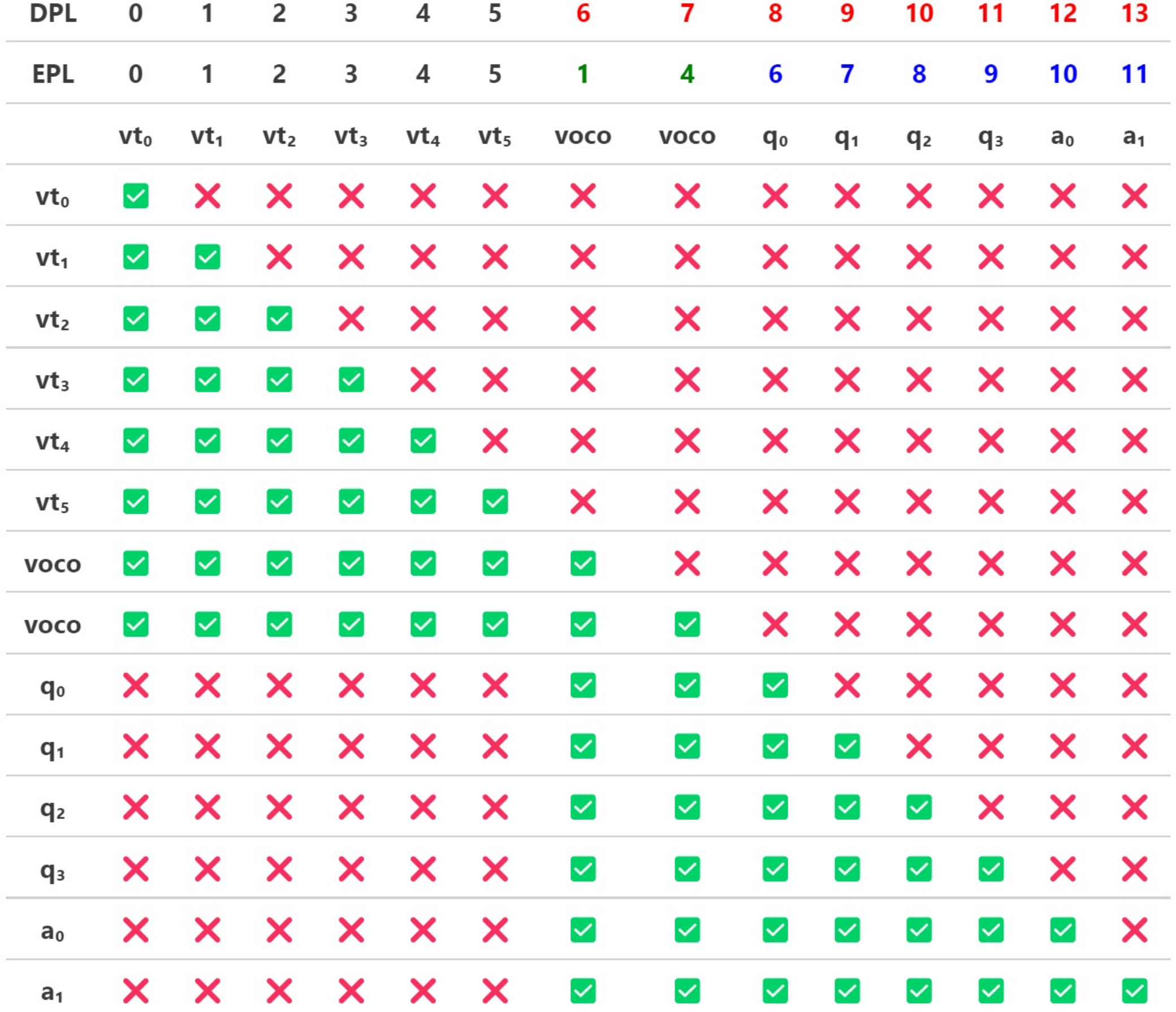}
\caption{Attention mask and position layout of VoCo-LLaMA~\cite{ye2025vocollamavisioncompressionlarge}. The Position IDs modified by EPL are marked in green (UPL) and blue (CPL) on top of the figure.}
\label{fig:voco_mask}
\end{figure*}

VoCo-LLaMA~\cite{ye2025vocollamavisioncompressionlarge} employs a modified attention mask which restricts text tokens from attending to vision tokens. Figure~\ref{fig:voco_mask} illustrates this mask, and we have indicated its position layout at the top of the figure.

\section{Attention Visualization}
\label{sec:attn_visualization_appedenx}
\begin{figure*}[htbp!]
\centering
\includegraphics[width=0.98\textwidth]{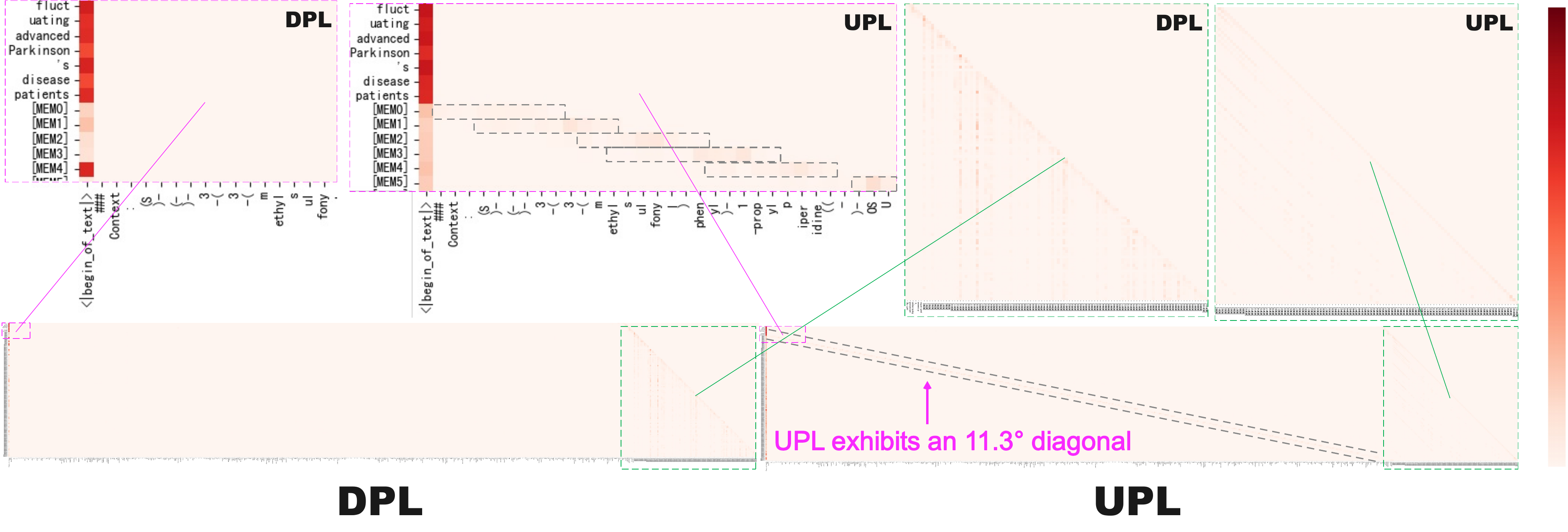}
\caption{Attention Matrix in the 2nd layer of ICAE after MRQA fine-tuning.}
\label{fig:layer_2_attn}
\end{figure*}

\begin{figure*}[htbp!]
\centering
\includegraphics[width=0.98\textwidth]{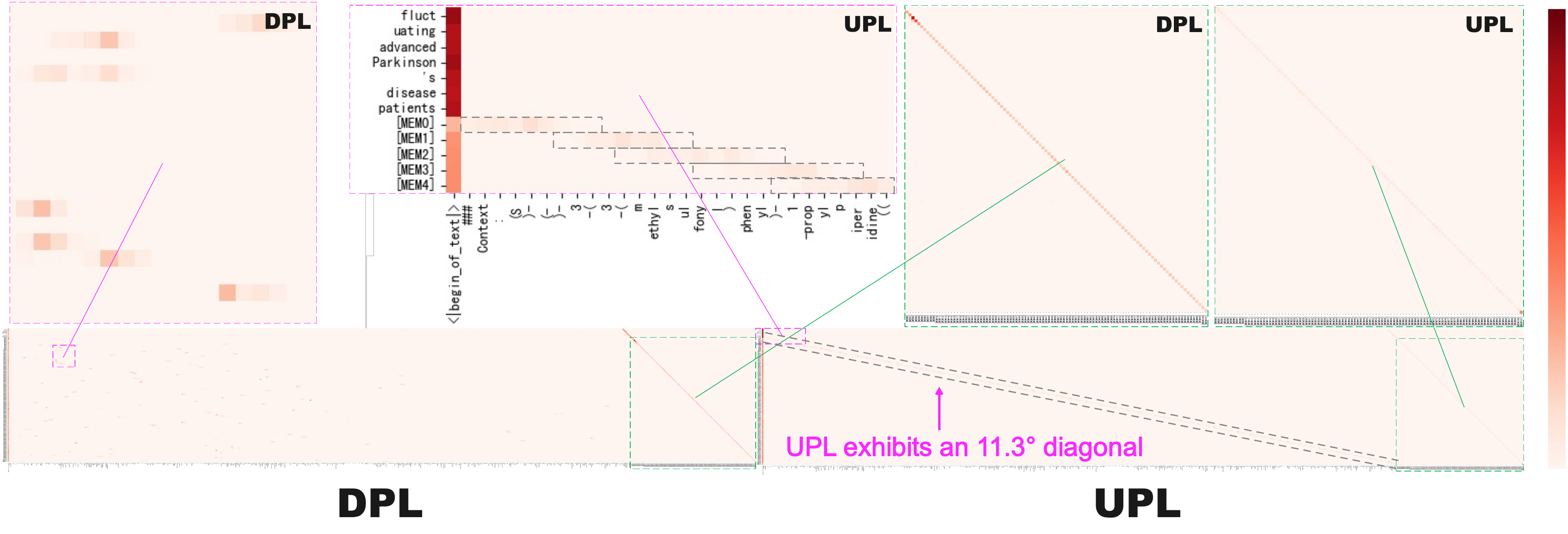}
\caption{Attention Matrix in the last layer of ICAE after MRQA fine-tuning.}
\label{fig:layer_last_attn}
\end{figure*}

Attention maps are presented in this section. Due to the low attention values of memory tokens in the first layer (which appear almost empty in the figure), we present the attention map of the second layer. See Figure~\ref{fig:layer_2_attn}. We provide a magnified view of the self-attention of memory tokens and their attention to other tokens. It can be observed that in the second layer, memory tokens in DPL only attend to themselves, while in UPL, they are able to attend to the entire context. Additionally, we also present the attention map of the last layer, shown in Figure~\ref{fig:layer_last_attn}. We use grey dashed lines to indicate the special attention pattern of UPL.

\section{Local bias of Position Encodings}
\label{sec:learn_pe_appendix}

\subsection{Sinusoidal Position Encoding}
\label{sec:sin_pe}
The sinusoidal position encoding~\cite{vaswani2023attentionneed,su2023roformerenhancedtransformerrotary} is given by:
$$
PE_{(pos, 2i)} = \sin\left(\frac{pos}{10000^{2i/d_{model}}}\right)
$$
$$
PE_{(pos, 2i+1)} = \cos\left(\frac{pos}{10000^{2i/d_{model}}}\right)
$$

Consider two nearby positions, $pos$ and $pos + \delta$, where $\delta$ is a small number.

For any dimension $i$, the argument to the sine/cosine function changes from $\frac{pos}{10000^{2i/d_{model}}}$ to $\frac{pos + \delta}{10000^{2i/d_{model}}}$.
The change in the argument is $\frac{\delta}{10000^{2i/d_{model}}}$.

For small $\delta$, this change in the argument is small for all dimensions.

Since sine and cosine functions are continuous, a small change in their input argument results in a small change in their output value.

$$
\sin(x + \epsilon) \approx \sin(x) \quad \text{for small } \epsilon
$$
$$
\cos(x + \epsilon) \approx \cos(x) \quad \text{for small } \epsilon
$$

Therefore, each dimension of $PE_{(pos+\delta)}$ is very close to the corresponding dimension of $PE_{(pos)}$.

This means the entire vector $PE_{(pos+\delta)}$ is very similar to $PE_{(pos)}$ when $\delta$ is small.

This property injects a local inductive bias.

\subsection{Learnable Position Encodings}
\begin{figure*}[htbp!]
\centering
\includegraphics[width=0.98\textwidth]{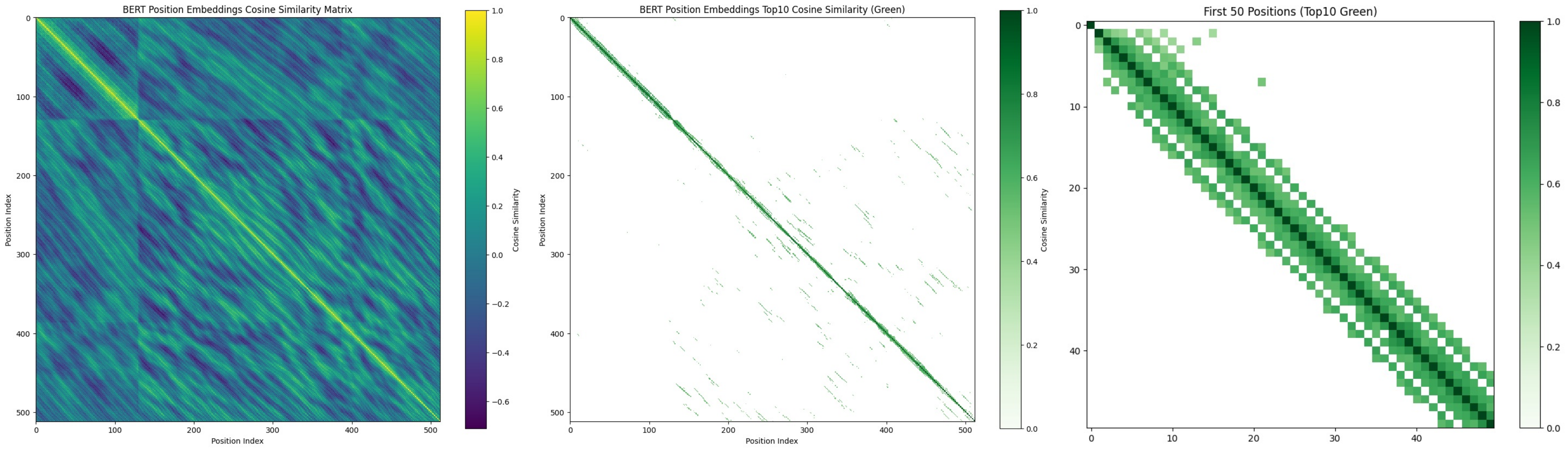}
\caption{The cosine similarity of BERT's positional encodings. To improve readability, we have removed results beyond the top 10 and zoomed in on the first 50 positions.
}
\label{fig:bert_pos}
\end{figure*}

\begin{figure*}[htbp!]
\centering
\includegraphics[width=0.98\textwidth]{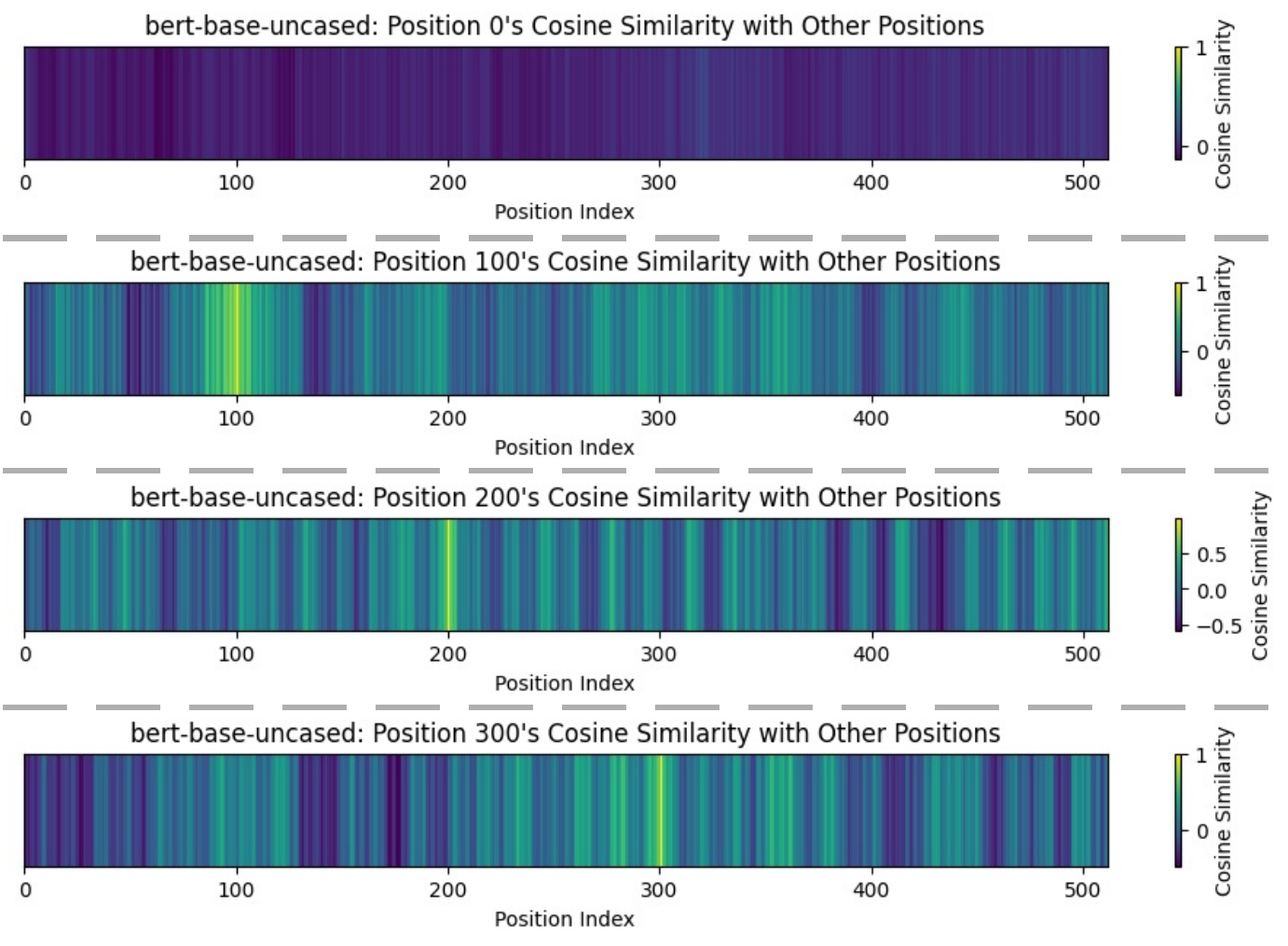}
\caption{The cosine similarity between the positional encoding of the \texttt{[CLS]} token and other positions.
}
\label{fig:bert_cls_pos}
\end{figure*}

In this section, we present figures related to learnable position embeddings. 
Figure~\ref{fig:bert_pos} illustrates the cosine similarity between different positions of BERT~\cite{devlin-etal-2019-bert}. It can be observed that the cosine similarity between nearby positions is significantly high. To illustrate the special behavior of the \texttt{[CLS]} token's position embedding, we show in Figure~\ref{fig:bert_cls_pos} the cosine similarity of position 0 (i.e., the position ID of [CLS]), position 100, position 200, and position 300 with other positions.
Figure~\ref{fig:t5_bias} illustrates that as the relative distance increases, the learnable bias added by T5~\cite{JMLR:v21:20-074} to the attention scores decreases.

\begin{table*}[htbp!]
  \centering
  \tiny
  \begin{tabular}{c|cccccccccc}
    \toprule
    & $x_{0}$ & $x_{1}$ & \dots & $x_{|X|-1}$ &  $y_{0}$ & $y_{1}$ & \dots & $y_{|Y|-1}$  \\
    \midrule
    \textbf{Default Position ID}     & $0$ & $1$ & \dots & $|X|-1$ &  $|X|$ & $|X|+1$ &  \dots & $|X|+|Y|-1$ \\
    \midrule
    \midrule
    & $KV(x_{0})$ & $KV(x_{1})$ & \dots & $KV(x_{|X|-1})$  & $y_{0}$ & $y_{1}$ & \dots & $y_{|Y|-1}$  \\
    \midrule
    \textbf{Default Position ID} & 0 & 1 & \dots & $|X|-1$ & $|X|$ & $|X| + 1$ &  \dots & $|X|+|Y|-1$ \\
    \bottomrule
  \end{tabular}
  \caption{
    Default position layout of Transformers.
  }
  \label{default-position-layout}
\end{table*}

\begin{table*}[htbp!]
  \centering
  \tiny
  \begin{tabular}{c|cccccccc}
    \toprule
    & $x_{(i-1)L}$ & $x_{(i-1)L+1}$ & \dots & $x_{iL-1}$ & $m_0$ & $m_1$ & \dots & $m_{|M|-1}$ \\
    \midrule
    \textbf{DPL} & 0 & 1 & \dots & $L-1$ & $L$ & $L+1$ & \dots & $L+|M|-1$ \\
    \textbf{EPL} & $(i-1)L+1$ & $(i-1)L+2$ & \dots & $iL$ & $\rd{b}$ & $\rd{b+r}$ & \dots & $\rd{(b+(|M|-1)r}$ \\
    \midrule
    \midrule

    \multicolumn{9}{c}{$i=1, L=510, |M|=102, r=5$} \\
    
    \midrule
    & $x_{0}$ & $x_{1}$ & \dots & $x_{509}$ & $m_0$ & $m_1$ & \dots & $m_{101}$ \\
    \midrule
    \textbf{DPL} & 0 & 1 & \dots & $509$ & $510$ & $511$ & \dots & $611$ \\
    \textbf{EPL} & $1$ & $2$ & \dots & $510$ & $3$ & $8$ & \dots & $508$ \\
    \midrule
    \midrule

    \multicolumn{9}{c}{$i=2, L=510, |M|=102, r=5$} \\
    
    \midrule
    & $x_{510}$ & $x_{511}$ & \dots & $x_{1019}$ & $m_0$ & $m_1$ & \dots & $m_{101}$ \\
    \midrule
    \textbf{DPL} & 0 & 1 & \dots & $509$ & $510$ & $511$ & \dots & $611$ \\
    \textbf{EPL} & $511$ & $512$ & \dots & $1020$ & $513$ & $518$ & \dots & $1018$ \\
    \bottomrule
  \end{tabular}
  \caption{Position Layout of Encoder for $S^{(i)}$. $r=\frac{L}{|M|}$; $b=(i-1)*L+1+\frac{r-1}{2}$. The notation $\rd{\cdot}$ indicates rounding to the nearest integer.} 
  \label{tab:encoder_layout_chunk2_detail}
\end{table*}


\begin{table*}[htbp!]
  \centering
  \tiny
  \begin{tabular}{c|ccccccccc}
    \toprule
    \noalign{\vspace{0.2em}}
    & $\tilde{m}_{0}^{(1)}/KV_{0}^{(1)}$ & $\tilde{m}_{1}^{(1)}/KV_{1}^{(1)}$ & \dots  & $\tilde{m}_{|M|-1}^{(k)}/KV_{|M|-1}^{(k)}$ & $\texttt{[AE]}$ & $x_{0}$ & $x_{1}$ & \dots & $x_{p-1}$ \\
    \noalign{\vspace{0.2em}}
    \midrule
    \textbf{DPL(ICAE)} & 0 & 1 & \dots & $k|M|-1$ & $k|M|$ & $k|M|+1$ & $k|M|+2$ & \dots & $k|M|+p$ \\
    \textbf{DPL(500x)} & $L$ & $L+1$ & \dots & $L+|M|-1$ & $k|M|$ & $k|M|+1$ & $k|M|+2$ & \dots & $k|M|+p$ \\
    \textbf{EPL} & $\rd{b}$ & $\rd{b+r}$  & \dots & $\rd{b+(k|M|-1)r}$ & $0$ & $1$ & $2$ & \dots & $p$ \\
    \midrule
    \midrule

    \multicolumn{10}{c}{$k=2, L=510, |M|=102, r=5, p=1020$} \\
    
    \midrule
    \noalign{\vspace{0.2em}}
    & $\tilde{m}_{0}^{(1)}/KV_{0}^{(1)}$ & $\tilde{m}_{1}^{(1)}/KV_{1}^{(1)}$ & \dots  & $\tilde{m}_{101}^{(2)}/KV_{101}^{(2)}$ & $\texttt{[AE]}$ & $x_{0}$ & $x_{1}$ & \dots & $x_{1019}$ \\
    \noalign{\vspace{0.2em}}
    \midrule
    \textbf{DPL(ICAE)} & $0$ & $1$ & \dots & $203$ & $204$ & $205$ & $206$ & \dots & $1224$ \\
    \textbf{DPL(500x)} & $510$ & $511$ & \dots & $611$ & $204$ & $205$ & $206$ & \dots & $1224$ \\
    \textbf{EPL} & $3$ & $8$  & \dots & $1018$ & $0$ & $1$ & $2$ & \dots & $1020$ \\
    \bottomrule

  \end{tabular}
  \caption{Position Layout of Decoder in AE Task. $r=\frac{L}{|M|}$; $b=1+\frac{r-1}{2}$. The notation $\rd{\cdot}$ indicates rounding to the nearest integer.}
  \label{tab:decoder_layout_ae_detail}
\end{table*}


\begin{table*}[htbp!]
  \centering
  \tiny
  \begin{tabular}{c|ccccccccc}
    \toprule
    \noalign{\vspace{0.2em}}
    & $\tilde{m}_{0}^{(1)}/KV_{0}^{(1)}$ & $\tilde{m}_{1}^{(1)}/KV_{1}^{(1)}$ & \dots  & $\tilde{m}_{|M|-1}^{(k)}/KV_{|M|-1}^{(k)}$ & $\texttt{[LM]}$ & $x_{p}$ & $x_{p+1}$ & \dots & $x_{|X|-1}$ \\
    \noalign{\vspace{0.2em}}
    \midrule
    \textbf{DPL(ICAE)} & 0 & 1 & \dots & $k|M|-1$ & $k|M|$ & $k|M|+1$ & $k|M|+2$ & \dots & $k|M|+|X|-p$ \\
    \textbf{DPL(500x)} & $L$ & $L+1$ & \dots & $L+|M|-1$ & $k|M|$ & $k|M|+1$ & $k|M|+2$ & \dots & $k|M|+|X|-p$ \\
    \textbf{EPL} & $\rd{b}$ & $\rd{b+r}$  & \dots & $\rd{b+(k|M|-1)r}$ & $p$ & $p+1$ & $p+2$ & \dots & $|X|$ \\
    \midrule
    \midrule

    \multicolumn{10}{c}{$k=2, L=510, |M|=102, r=5, |X|=2040,p=1020$} \\
    
    \midrule
    \noalign{\vspace{0.2em}}
     & $\tilde{m}_{0}^{(1)}/KV_{0}^{(1)}$ & $\tilde{m}_{1}^{(1)}/KV_{1}^{(1)}$ & \dots  & $\tilde{m}_{101}^{(2)}/KV_{101}^{(2)}$ & $\texttt{[LM]}$ & $x_{1020}$ & $x_{1021}$ & \dots & $x_{2039}$ \\
    \noalign{\vspace{0.2em}}
    \midrule
    \textbf{DPL(ICAE)} & $0$ & $1$ & \dots & $203$ & $204$ & $205$ & $206$ & \dots & $1224$ \\
    \textbf{DPL(500x)} & $510$ & $511$ & \dots & $611$ & $204$ & $205$ & $206$ & \dots & $1224$ \\
    \textbf{EPL} & $3$ & $8$  & \dots & $1018$ & $1020$ & $1021$ & $1022$ & \dots & $2040$ \\
    \bottomrule

  \end{tabular}
  \caption{Position Layout of Decoder in LM Task. $r=\frac{L}{|M|}$; $b=1+\frac{r-1}{2}$. The notation $\rd{\cdot}$ indicates rounding to the nearest integer.}
  \label{tab:decoder_layout_lm_detail}
\end{table*}


\begin{table*}[htbp!]
  \centering
  \tiny
  \begin{tabular}{c|ccccccccccc}
    \toprule
    \noalign{\vspace{0.2em}}
    & $\tilde{m}_{0}^{(1)}/KV_{0}^{(1)}$ & $\tilde{m}_{1}^{(1)}/KV_{1}^{(1)}$ & \dots  & $\tilde{m}_{|M|-1}^{(k)}/KV_{|M|-1}^{(k)}$ & $\texttt{[LM]}$ & $q_{0}$ & \dots & $q_{|Q|-1}$ & $a_{0}$ & \dots & $a_{|A|-1}$ \\
    \noalign{\vspace{0.2em}}
    \midrule
    \textbf{DPL(ICAE)} & 0 & 1 & \dots & $k|M|-1$ & $k|M|$ & $k|M|+1$ & \dots & $t$ & $t+1$ & \dots & $t+|A|$\\
    \textbf{DPL(500x)} & $L$ & $L+1$ & \dots & $L+|M|-1$ & $k|M|$ & $k|M|+1$ & \dots & $t$ & $t+1$ & \dots & $t+|A|$\\
    \textbf{EPL} & $\rd{b}$ & $\rd{b+r}$  & \dots & $\rd{b+(k|M|-1)r}$ & $|C|$ & $|C|+1$ & \dots & $t'$ & $t'+1$ & \dots & $t'+|A|$\\
    \midrule
    \midrule

    \multicolumn{12}{c}{$k=2, L=510, |M|=102, r=5, |C|=1020,|Q|=50,|A|=5$} \\
    
    \midrule
    \noalign{\vspace{0.2em}}
     & $\tilde{m}_{0}^{(1)}/KV_{0}^{(1)}$ & $\tilde{m}_{1}^{(1)}/KV_{1}^{(1)}$ & \dots  & $\tilde{m}_{101}^{(2)}/KV_{101}^{(2)}$ & $\texttt{[LM]}$ & $q_{0}$ & \dots & $q_{49}$ & $a_{0}$ & \dots & $a_{4}$ \\
    \noalign{\vspace{0.2em}}
    \midrule
    \textbf{DPL(ICAE)} & $0$ & $1$ & \dots & $203$ & $204$ & $205$ & \dots & $254$ & $255$ & \dots & $259$ \\
    \textbf{DPL(500x)} & $510$ & $511$ & \dots & $611$ & $204$ & $205$ & \dots & $254$ & $255$ & \dots & $259$ \\
    \textbf{EPL} & $3$ & $8$  & \dots & $1018$ & $1020$ & $1021$ & \dots & $1070$ & $1071$ & \dots & $1075$ \\
    \bottomrule

  \end{tabular}
  \caption{Position Layout of Decoder in QA Task. $r=\frac{L}{|M|}$; $b=1+\frac{r-1}{2}$; $t=k|M|+|Q|$; $t'=|C|+|Q|$. The notation $\rd{\cdot}$ indicates rounding to the nearest integer.}
  \label{tab:decoder_layout_qa}
\end{table*}


\begin{table*}[htbp!]
    \centering
    \tiny
    \begin{tabular}{l c c c c c c c c c c c c c c}
        \toprule
        & \multicolumn{2}{c}{\textbf{SQuAD}} & \multicolumn{2}{c}{\textbf{NewsQA}} & \multicolumn{2}{c}{\textbf{TriQA}} & \multicolumn{2}{c}{\textbf{SearchQA}} & \multicolumn{2}{c}{\textbf{HQA}} & \multicolumn{2}{c}{\textbf{NQ}} & \multicolumn{2}{c}{\textbf{Avg.}} \\
        \cmidrule(lr){2-3} \cmidrule(lr){4-5} \cmidrule(lr){6-7} \cmidrule(lr){8-9} \cmidrule(lr){10-11} \cmidrule(lr){12-13} \cmidrule(lr){14-15}
        & F1 & EM & F1 & EM & F1 & EM & F1 & EM & F1 & EM & F1 & EM & F1 & EM \\
        \midrule
        \multicolumn{15}{l}{\textbf{Llama-3.2-1B}} \\
        \midrule
        \textbf{NoContext} & \textcolor{SoftRed}{9.34} & \textcolor{SoftRed}{1.92} & \textcolor{SoftRed}{4.80} & \textcolor{SoftRed}{0.59} & \textcolor{SoftRed}{3.04} & \textcolor{SoftRed}{0.92} & \textcolor{SoftRed}{14.99} & \textcolor{SoftRed}{8.71} & \textcolor{SoftRed}{9.47} & \textcolor{SoftRed}{3.27} & \textcolor{SoftRed}{10.88} & \textcolor{SoftRed}{3.91} & \textcolor{SoftRed}{8.75} & \textcolor{SoftRed}{3.22} \\
        ICAE(DPL) & 54.58 & 36.99 & 37.57 & 21.25 & 57.96 & 48.85 & 68.90 & 56.68 & 59.61 & 43.01 & 59.40 & 42.01 & 56.34 & 41.47 \\
        ICAE(EPL) & \best{59.94} & \best{39.39} & \best{43.61} & \best{24.95} & \best{61.50} & \best{51.98} & \best{69.31} & \best{57.11} & \best{64.22} & \best{46.84} & \best{61.21} & \best{42.11} & \best{59.97} & \best{43.73} \\

        500x(DPL) & \best{68.26} & \best{49.15} & 43.17 & 25.31 & 60.30 & 51.03 & 70.33 & 57.96 & 66.44 & 49.45 & 64.69 & 46.78 & 62.20 & 46.61 \\
        500x(EPL) & 67.87 & 47.54 & \best{49.23} & \best{29.58} & \best{64.53} & \best{55.16} & \best{72.51} & \best{60.39} & \best{68.51} & \best{51.06} & \best{65.55} & \best{46.97} & \best{64.70} & \best{48.45} \\

        \textbf{FullContext} & \textcolor{SoftGreen}{58.72} & \textcolor{SoftGreen}{39.20} & \textcolor{SoftGreen}{38.72} & \textcolor{SoftGreen}{16.62} & \textcolor{SoftGreen}{31.82} & \textcolor{SoftGreen}{24.42} & \textcolor{SoftGreen}{49.15} & \textcolor{SoftGreen}{36.27} & \textcolor{SoftGreen}{53.22} & \textcolor{SoftGreen}{39.08} & \textcolor{SoftGreen}{53.47} & \textcolor{SoftGreen}{36.12} & \textcolor{SoftGreen}{47.51} & \textcolor{SoftGreen}{31.95} \\
        \midrule
        \multicolumn{15}{l}{\textbf{Llama-3.2-3B}} \\
        \midrule
        \textbf{NoContext} & \textcolor{SoftRed}{15.65} & \textcolor{SoftRed}{7.09} & \textcolor{SoftRed}{6.15} & \textcolor{SoftRed}{1.26} & \textcolor{SoftRed}{8.68} & \textcolor{SoftRed}{6.38} & \textcolor{SoftRed}{42.54} & \textcolor{SoftRed}{31.83} & \textcolor{SoftRed}{16.31} & \textcolor{SoftRed}{9.02} & \textcolor{SoftRed}{16.43} & \textcolor{SoftRed}{8.18} & \textcolor{SoftRed}{17.63} & \textcolor{SoftRed}{10.63} \\
        ICAE(DPL) & 58.95 & 41.53 & 39.70 & 23.12 & 60.49 & 51.64 & 66.35 & 53.81 & 58.68 & 42.55 & 58.55 & 41.94 & 57.12 & 42.43 \\
        ICAE(EPL) & \best{73.82} & \best{53.67} & \best{58.02} & \best{36.51} & \best{71.22} & \best{62.26} & \best{71.41} & \best{59.90} & \best{72.61} & \best{55.77} & \best{70.19} & \best{51.33} & \best{69.55} & \best{53.24} \\
        500x(DPL) & 68.28 & 50.15 & 50.98 & 32.15 & 68.10 & 59.54 & 74.16 & 62.11 & 70.63 & 53.57 & 67.58 & 50.19 & 66.62 & 51.28 \\
        500x(EPL) & \best{77.74} & \best{58.38} & \best{61.19} & \best{41.62} & \best{71.62} & \best{62.66} & \best{74.32} & \best{62.67} & \best{75.06} & \best{58.52} & \best{72.03} & \best{54.07} & \best{71.99} & \best{56.32} \\

        \textbf{FullContext} & \textcolor{SoftGreen}{74.07} & \textcolor{SoftGreen}{55.43} & \textcolor{SoftGreen}{48.99} & \textcolor{SoftGreen}{24.62} & \textcolor{SoftGreen}{63.92} & \textcolor{SoftGreen}{54.05} & \textcolor{SoftGreen}{67.69} & \textcolor{SoftGreen}{52.31} & \textcolor{SoftGreen}{63.73} & \textcolor{SoftGreen}{48.33} & \textcolor{SoftGreen}{64.32} & \textcolor{SoftGreen}{46.54} & \textcolor{SoftGreen}{63.79} & \textcolor{SoftGreen}{46.88} \\
        \midrule
        \multicolumn{15}{l}{\textbf{Llama-3.1-8B}} \\
        \midrule
        \textbf{NoContext} & \textcolor{SoftRed}{21.84} & \textcolor{SoftRed}{11.74} & \textcolor{SoftRed}{9.52} & \textcolor{SoftRed}{3.28} & \textcolor{SoftRed}{31.58} & \textcolor{SoftRed}{26.37} & \textcolor{SoftRed}{59.66} & \textcolor{SoftRed}{45.49} & \textcolor{SoftRed}{20.34} & \textcolor{SoftRed}{12.71} & \textcolor{SoftRed}{29.62} & \textcolor{SoftRed}{17.94} & \textcolor{SoftRed}{28.76} & \textcolor{SoftRed}{19.59} \\
        ICAE(DPL) & 56.56 & 38.87 & 36.99 & 20.06 & 64.54 & 55.58 & 71.35 & 58.92 & 57.04 & 41.20 & 58.04 & 41.19 & 57.42 & 42.64 \\
        ICAE(EPL) & \best{78.44} & \best{58.90} & \best{61.69} & \best{40.17} & \best{73.92} & \best{65.05} & \best{80.05} & \best{67.65} & \best{74.93} & \best{58.33} & \best{72.43} & \best{54.14} & \best{73.58} & \best{57.37} \\

        500x(DPL) & 80.56 & \best{62.11} & 60.31 & 40.65 & 74.00 & 65.39 & \best{79.60} & 67.51 & 76.18 & 59.62 & 74.17 & \best{56.48} & 74.14 & 58.63 \\
        500x(EPL) & \best{80.60} & 61.37 & \best{64.43} & \best{44.35} & \best{74.75} & \best{66.04} & 79.39 & \best{67.74} & \best{77.17} & \best{60.53} & \best{74.71} & 56.16 & \best{75.18} & \best{59.36} \\
        \textbf{FullContext} & \textcolor{SoftGreen}{80.53} & \textcolor{SoftGreen}{61.97} & \textcolor{SoftGreen}{60.24} & \textcolor{SoftGreen}{40.05} & \textcolor{SoftGreen}{72.65} & \textcolor{SoftGreen}{63.28} & \textcolor{SoftGreen}{76.54} & \textcolor{SoftGreen}{61.99} & \textcolor{SoftGreen}{73.07} & \textcolor{SoftGreen}{57.19} & \textcolor{SoftGreen}{72.25} & \textcolor{SoftGreen}{54.01} & \textcolor{SoftGreen}{72.55} & \textcolor{SoftGreen}{56.42} \\
        \bottomrule
    \end{tabular}
    \caption{Results on the in-domain validation set, including six QA datasets: SQuAD~\cite{rajpurkar-etal-2016-squad}, NewsQA~\cite{trischler-etal-2017-newsqa}, TriviaQA (TriQA)~\cite{joshi-etal-2017-triviaqa}, SearchQA~\cite{DBLP:journals/corr/DunnSHGCC17}, HotpotQA (HQA)~\cite{yang-etal-2018-hotpotqa}, and NaturalQuestions (NQ)~\cite{47761}.}
    \label{tab:id_results}
\end{table*}


\begin{table*}[htbp!]
    \centering
    \tiny
    \begin{tabular}{l c c c c c c c c c c c c c c}
        \toprule
        & \multicolumn{2}{c}{\textbf{BioASQ}} & \multicolumn{2}{c}{\textbf{DROP}} & \multicolumn{2}{c}{\textbf{DuoRC}} & \multicolumn{2}{c}{\textbf{RACE}} & \multicolumn{2}{c}{\textbf{RE}} & \multicolumn{2}{c}{\textbf{TQA}} & \multicolumn{2}{c}{\textbf{Avg.}} \\
        \cmidrule(lr){2-3} \cmidrule(lr){4-5} \cmidrule(lr){6-7} \cmidrule(lr){8-9} \cmidrule(lr){10-11} \cmidrule(lr){12-13} \cmidrule(lr){14-15}
        & F1 & EM & F1 & EM & F1 & EM & F1 & EM & F1 & EM & F1 & EM & F1 & EM \\
        \midrule
        \multicolumn{15}{l}{\textbf{Llama-3.2-1B}} \\
        \midrule
        \textbf{NoContext} & \textcolor{SoftRed}{10.63} & \textcolor{SoftRed}{4.52} & \textcolor{SoftRed}{17.72} & \textcolor{SoftRed}{10.18} & \textcolor{SoftRed}{3.93} & \textcolor{SoftRed}{0.53} & \textcolor{SoftRed}{6.58} & \textcolor{SoftRed}{0.30} & \textcolor{SoftRed}{11.74} & \textcolor{SoftRed}{3.70} & \textcolor{SoftRed}{20.50} & \textcolor{SoftRed}{9.78} & \textcolor{SoftRed}{11.85} & \textcolor{SoftRed}{4.83} \\
        ICAE(DPL) & 38.39 & 26.99 & \best{37.13} & \best{27.28} & 27.90 & 18.32 & 21.67 & 4.30 & \best{76.98} & \best{64.45} & 37.62 & 22.55 & 39.95 & 27.32 \\
        ICAE(EPL) & \best{42.66} & \best{29.19} & 37.03 & 26.95 & \best{32.50} & \best{21.39} & \best{26.87} & \best{5.34} & 76.90 & 64.31 & \best{47.25} & \best{29.81} & \best{43.87} & \best{29.50} \\
        500x(DPL) & 44.22 & 31.38 & 40.47 & 29.34 & 32.20 & 21.19 & 27.67 & 5.93 & \best{83.03} & \best{71.98} & 47.00 & 28.88 & 45.76 & 31.45 \\
        500x(EPL) & \best{46.11} & \best{31.91} & \best{40.94} & \best{29.41} & \best{39.07} & \best{26.32} & \best{31.89} & \best{7.27} & 80.39 & 67.71 & \best{49.80} & \best{31.14} & \best{48.03} & \best{32.29} \\
        \textbf{FullContext} & \textcolor{SoftGreen}{45.72} & \textcolor{SoftGreen}{31.45} & \textcolor{SoftGreen}{40.74} & \textcolor{SoftGreen}{30.27} & \textcolor{SoftGreen}{39.20} & \textcolor{SoftGreen}{27.85} & \textcolor{SoftGreen}{29.57} & \textcolor{SoftGreen}{7.72} & \textcolor{SoftGreen}{74.85} & \textcolor{SoftGreen}{64.01} & \textcolor{SoftGreen}{53.31} & \textcolor{SoftGreen}{34.60} & \textcolor{SoftGreen}{47.23} & \textcolor{SoftGreen}{32.65} \\
        \midrule
        \multicolumn{15}{l}{\textbf{Llama-3.2-3B}} \\
        \midrule
        \textbf{NoContext} & \textcolor{SoftRed}{24.84} & \textcolor{SoftRed}{17.22} & \textcolor{SoftRed}{21.17} & \textcolor{SoftRed}{13.37} & \textcolor{SoftRed}{5.87} & \textcolor{SoftRed}{1.93} & \textcolor{SoftRed}{8.87} & \textcolor{SoftRed}{1.19} & \textcolor{SoftRed}{20.60} & \textcolor{SoftRed}{13.30} & \textcolor{SoftRed}{37.34} & \textcolor{SoftRed}{22.82} & \textcolor{SoftRed}{19.78} & \textcolor{SoftRed}{11.64} \\
        ICAE(DPL) & 41.75 & 31.65 & 38.77 & 29.81 & 28.40 & 18.99 & 21.58 & 3.56 & 76.56 & 64.82 & 47.85 & 30.41 & 42.49 & 29.87 \\
        ICAE(EPL) & \best{50.83} & \best{36.70} & \best{54.10} & \best{43.38} & \best{45.66} & \best{33.24} & \best{38.78} & \best{9.35} & \best{82.71} & \best{71.64} & \best{58.13} & \best{36.93} & \best{55.03} & \best{38.54} \\
        500x(DPL) & 50.20 & 37.63 & 48.64 & 38.32 & 38.77 & 26.85 & 31.86 & 8.16 & 83.67 & 73.71 & 55.08 & 34.80 & 51.37 & 36.58 \\
        500x(EPL) & \best{53.15} & \best{39.03} & \best{57.00} & \best{45.84} & \best{50.35} & \best{36.51} & \best{42.45} & \best{10.83} & \best{85.20} & \best{75.20} & \best{58.09} & \best{35.53} & \best{57.71} & \best{40.49} \\
        \textbf{FullContext} & \textcolor{SoftGreen}{59.20} & \textcolor{SoftGreen}{42.82} & \textcolor{SoftGreen}{50.30} & \textcolor{SoftGreen}{35.53} & \textcolor{SoftGreen}{39.09} & \textcolor{SoftGreen}{27.51} & \textcolor{SoftGreen}{36.72} & \textcolor{SoftGreen}{9.20} & \textcolor{SoftGreen}{81.73} & \textcolor{SoftGreen}{73.27} & \textcolor{SoftGreen}{67.50} & \textcolor{SoftGreen}{43.65} & \textcolor{SoftGreen}{55.76} & \textcolor{SoftGreen}{38.66} \\
        \midrule
        \multicolumn{15}{l}{\textbf{Llama-3.2-8B}} \\
        \midrule
        \textbf{NoContext} & \textcolor{SoftRed}{43.16} & \textcolor{SoftRed}{33.64} & \textcolor{SoftRed}{25.60} & \textcolor{SoftRed}{18.43} & \textcolor{SoftRed}{6.15} & \textcolor{SoftRed}{2.33} & \textcolor{SoftRed}{7.98} & \textcolor{SoftRed}{1.63} & \textcolor{SoftRed}{33.89} & \textcolor{SoftRed}{25.27} & \textcolor{SoftRed}{50.27} & \textcolor{SoftRed}{32.67} & \textcolor{SoftRed}{27.84} & \textcolor{SoftRed}{19.00} \\
        ICAE(DPL) & 47.42 & 33.98 & 35.98 & 27.41 & 22.54 & 13.99 & 21.48 & 4.60 & 77.58 & 65.98 & 53.55 & 34.13 & 43.09 & 30.01 \\
        ICAE(EPL) & \best{53.21} & \best{37.90} & \best{58.83} & \best{47.50} & \best{47.26} & \best{33.44} & \best{40.65} & \best{9.64} & \best{84.35} & \best{73.91} & \best{58.39} & \best{36.13} & \best{57.12} & \best{39.76} \\
        500x(DPL) & 51.50 & 37.70 & 59.85 & 47.64 & 47.95 & 34.44 & 41.17 & 10.39 & \best{85.86} & \best{75.85} & \best{58.22} & \best{35.93} & 57.42 & 40.32 \\
        500x(EPL) & \best{54.80} & \best{39.43} & \best{62.46} & \best{50.70} & \best{51.33} & \best{37.38} & \best{43.40} & \best{11.13} & 85.77 & 75.10 & 58.11 & 35.00 & \best{59.31} & \best{41.45} \\
        \textbf{FullContext} & \textcolor{SoftGreen}{59.80} & \textcolor{SoftGreen}{42.95} & \textcolor{SoftGreen}{60.64} & \textcolor{SoftGreen}{47.64} & \textcolor{SoftGreen}{26.91} & \textcolor{SoftGreen}{16.86} & \textcolor{SoftGreen}{42.80} & \textcolor{SoftGreen}{10.53} & \textcolor{SoftGreen}{85.11} & \textcolor{SoftGreen}{74.08} & \textcolor{SoftGreen}{71.68} & \textcolor{SoftGreen}{47.57} & \textcolor{SoftGreen}{57.83} & \textcolor{SoftGreen}{39.94} \\
        \bottomrule
    \end{tabular}
    \caption{Results on the out-of-domain validation set, including six QA datasets: BioASQ~\cite{Tsatsaronis2015}, DROP~\cite{dua-etal-2019-drop}, DuoRC~\cite{saha-etal-2018-duorc}, RACE~\cite{lai-etal-2017-race}, Relation Extraction (RE)~\cite{levy-etal-2017-zero}, and TextbookQA (TQA)~\cite{Kembhavi_2017_CVPR}.}
    \label{tab:ood_results}
\end{table*}


\begin{table*}[htbp!]
    \centering
    \tiny
    \begin{tabular}{l c c c c c c c c c c c c c c}
        \toprule
        & \multicolumn{2}{c}{\textbf{SQuAD}} & \multicolumn{2}{c}{\textbf{NewsQA}} & \multicolumn{2}{c}{\textbf{TriQA}} & \multicolumn{2}{c}{\textbf{SearchQA}} & \multicolumn{2}{c}{\textbf{HQA}} & \multicolumn{2}{c}{\textbf{NQ}} & \multicolumn{2}{c}{\textbf{Avg.}} \\
        \cmidrule(lr){2-3} \cmidrule(lr){4-5} \cmidrule(lr){6-7} \cmidrule(lr){8-9} \cmidrule(lr){10-11} \cmidrule(lr){12-13} \cmidrule(lr){14-15}
        & F1 & EM & F1 & EM & F1 & EM & F1 & EM & F1 & EM & F1 & EM & F1 & EM \\
        \midrule
        \multicolumn{15}{l}{\textbf{Llama-3.2-1B}} \\
        \midrule
        ICAE(DPL) & 54.58 & 36.99 & 37.57 & 21.25 & 57.96 & 48.85 & 68.90 & 56.68 & 59.61 & 43.01 & 59.40 & 42.01 & 56.34 & 41.47 \\
        ICAE(UPL) & \best{62.22} & \best{42.25} & \best{44.49} & \best{26.16} & 61.35 & 51.88 & \best{70.33} & \best{57.73} & 63.85 & \best{46.89} & \best{61.71} & \best{43.00} & \best{60.66} & \best{44.65} \\
        ICAE(EPL) & 59.94 & 39.39 & 43.61 & 24.95 & \best{61.50} & \best{51.98} & 69.31 & 57.11 & \best{64.22} & 46.84 & 61.21 & 42.11 & 59.97 & 43.73 \\

        500x(DPL) & \best{68.26} & \best{49.15} & 43.17 & 25.31 & 60.30 & 51.03 & 70.33 & 57.96 & 66.44 & 49.45 & 64.69 & 46.78 & 62.20 & 46.61 \\
        500x(UPL) & 65.92 & 46.35 & \best{49.33} & \best{29.89} & 63.97 & 54.98 & 71.85 & 59.77 & 67.04 & 50.53 & 64.20 & 45.66 & 63.72 & 47.86 \\
        500x(EPL) & 67.87 & 47.54 & 49.23 & 29.58 & \best{64.53} & \best{55.16} & \best{72.51} & \best{60.39} & \best{68.51} & \best{51.06} & \best{65.55} & \best{46.97} & \best{64.70} & \best{48.45} \\

        \bottomrule
    \end{tabular}
    \caption{Ablation results on the in-domain validation set, including six QA datasets: SQuAD~\cite{rajpurkar-etal-2016-squad}, NewsQA~\cite{trischler-etal-2017-newsqa}, TriviaQA (TriQA)~\cite{joshi-etal-2017-triviaqa}, SearchQA~\cite{DBLP:journals/corr/DunnSHGCC17}, HotpotQA (HQA)~\cite{yang-etal-2018-hotpotqa}, and NaturalQuestions (NQ)~\cite{47761}.}
    \label{tab:ablation_id_results}
\end{table*}


\begin{table*}[htbp!]
    \centering
    \tiny
    \begin{tabular}{l c c c c c c c c c c c c c c}
        \toprule
        & \multicolumn{2}{c}{\textbf{BioASQ}} & \multicolumn{2}{c}{\textbf{DROP}} & \multicolumn{2}{c}{\textbf{DuoRC}} & \multicolumn{2}{c}{\textbf{RACE}} & \multicolumn{2}{c}{\textbf{RE}} & \multicolumn{2}{c}{\textbf{TQA}} & \multicolumn{2}{c}{\textbf{Avg.}} \\
        \cmidrule(lr){2-3} \cmidrule(lr){4-5} \cmidrule(lr){6-7} \cmidrule(lr){8-9} \cmidrule(lr){10-11} \cmidrule(lr){12-13} \cmidrule(lr){14-15}
        & F1 & EM & F1 & EM & F1 & EM & F1 & EM & F1 & EM & F1 & EM & F1 & EM \\
        \midrule
        \multicolumn{15}{l}{\textbf{Llama-3.2-1B}} \\
        \midrule
        ICAE(DPL) & 38.39 & 26.99 & 37.13 & 27.28 & 27.90 & 18.32 & 21.67 & 4.30 & \best{76.98} & \best{64.45} & 37.62 & 22.55 & 39.95 & 27.32 \\
        ICAE(UPL) & 41.82 & \best{29.32} & \best{37.77} & \best{27.41} & \best{33.73} & \best{22.45} & \best{26.90} & 4.90 & 73.61 & 60.62 & \best{48.19} & \best{29.94} & 43.67 & 29.11 \\
        ICAE(EPL) & \best{42.66} & 29.19 & 37.03 & 26.95 & 32.50 & 21.39 & 26.87 & \best{5.34} & 76.90 & 64.31 & 47.25 & 29.81 & \best{43.87} & \best{29.50} \\
        500x(DPL) & 44.22 & 31.38 & 40.47 & 29.34 & 32.20 & 21.19 & 27.67 & 5.93 & \best{83.03} & \best{71.98} & 47.00 & 28.88 & 45.76 & 31.45 \\
        500x(UPL) & 45.71 & \best{32.38} & 40.19 & 28.81 & 37.66 & 26.12 & 29.02 & 6.08 & 78.22 & 64.55 & \best{50.19} & \best{31.87} & 46.83 & 31.64 \\
        500x(EPL) & \best{46.11} & 31.91 & \best{40.94} & \best{29.41} & \best{39.07} & \best{26.32} & \best{31.89} & \best{7.27} & 80.39 & 67.71 & 49.80 & 31.14 & \best{48.03} & \best{32.29} \\
        \bottomrule
    \end{tabular}
    \caption{Ablation results on the out-of-domain validation set, including six QA datasets: BioASQ~\cite{Tsatsaronis2015}, DROP~\cite{dua-etal-2019-drop}, DuoRC~\cite{saha-etal-2018-duorc}, RACE~\cite{lai-etal-2017-race}, Relation Extraction (RE)~\cite{levy-etal-2017-zero}, and TextbookQA (TQA)~\cite{Kembhavi_2017_CVPR}.}
    \label{tab:ablation_ood_results}
\end{table*}


\end{document}